\documentclass[letterpaper, 10 pt, conference]{ieeeconf}  

\IEEEoverridecommandlockouts                              

\usepackage{times} 

\usepackage{float}
\usepackage{newtxmath}        
\usepackage{graphicx}
\usepackage{hyperref}

\title{\LARGE \bf
Geometry-aided Vision-based Localization of Future Mars Helicopters in Challenging Illumination Conditions
}

\author{
Dario Pisanti$^{1,*}$,
Robert Hewitt$^{2}$,
Roland Brockers$^{2}$,
Georgios Georgakis$^{2}$%
\thanks{$^{1}$Space Robotics Research Group, SnT, University of Luxembourg.}%
\thanks{$^{2}$Jet Propulsion Laboratory, California Institute of Technology.}%
\thanks{$^{*}$Corresponding author: dario.pisanti@uni.lu.}%
}

\begin{document}

\maketitle
\thispagestyle{empty}
\pagestyle{empty}

\begin{abstract}

Planetary exploration using aerial assets has the potential for unprecedented scientific discoveries on Mars. While NASA's Mars helicopter Ingenuity proved flight in Martian atmosphere is possible, future Mars rotorcraft will require advanced navigation capabilities for long-range flights. One such critical capability is Map-based Localization (MbL) which registers an onboard image to a reference map during flight to mitigate cumulative drift from visual odometry. However, significant illumination differences between rotorcraft observations and a reference map prove challenging for traditional MbL systems, restricting the operational window of the vehicle. 
In this work, we investigate a new MbL system and propose Geo-LoFTR, a geometry-aided deep learning model for image registration that is more robust under large illumination differences than prior models. The system is supported by a custom simulation framework that uses real orbital maps to produce large amounts of realistic images of the Martian terrain. Comprehensive evaluations show that our proposed system outperforms prior MbL efforts in terms of localization accuracy under significant lighting and scale variations. Furthermore, we demonstrate the validity of our approach across a simulated Martian day and on real Mars imagery. Code and datasets are available at: \href{https://dpisanti.github.io/geo-loftr/}{https://dpisanti.github.io/geo-loftr/}.

\end{abstract}


\section{INTRODUCTION}
\label{sec:intro}

The demonstration flights of NASA's Mars Helicopter, \textit{Ingenuity}, have marked a groundbreaking milestone in Mars exploration~\cite{Grip2022}. One of the next evolutionary steps is the Mars Science Helicopter (MSH)~\cite{Tzanetos2022}, a hexacopter concept with multi-kg payload capacity and 10 km traverse capability at altitudes up to 100 m, aimed at supporting high-priority investigations in Martian astrobiology, climate, and geology~\cite{Bapst2021Mars}. Precision navigation during such long-range traverses requires minimizing drift in position estimates generated by the onboard Visual Inertial Odometry (VIO) in a global navigation satellite system (GNSS)-denied environment like Mars. During flight demonstrations on Mars, Ingenuity's VIO exhibited 2-6\% position drift across traverses up to 625 m. These errors are expected to be considerably higher for MSH, requiring online global localization. A rotorcraft can be geo-localized by registering images captured by its navigation camera onto orbital maps pre-registered to a global reference frame, and estimating its pose relative to the map. This inherently drift-free geo-localization technique is referred to as Map-based Localization (MbL).

At the heart of every MbL system lies an image registration method that identifies distinctive features in the onboard image and the map to enable localization. This process is challenging due to large differences in lighting and scale between onboard imagery and the reference map. High-relief terrain may strengthen visual disparity at different times of day due to changes in shadow casting, while textureless regions can hinder the identification of distinctive features.

In spite of recent progress of deep learning in visual tasks, space applications still rely mostly on template-matching techniques or hand-crafted features to solve a relatively narrow registration problem with strong assumptions on viewpoint, scale, and lighting conditions. However, such assumptions limit the operational capability of future Mars rotorcraft. For example, the lack of robustness to varying illumination would restrict the flights to the time of day when the orbital map was originally collected, thus constraining the operational mission envelop. While recent deep learning methods~\cite{loftr,roma} have demonstrated robustness to illumination and scale variations on in-the-wild datasets such as MegaDepth~\cite{megadepth} for terrestrial applications, the main bottleneck is the lack of large-scale datasets that would allow finetuning these methods in planetary domains.
\begin{figure*}
    \vspace{3pt}
    \setlength{\abovecaptionskip}{0pt}
    \setlength{\belowcaptionskip}{0pt}
    \centering
    \includegraphics[width=\linewidth]{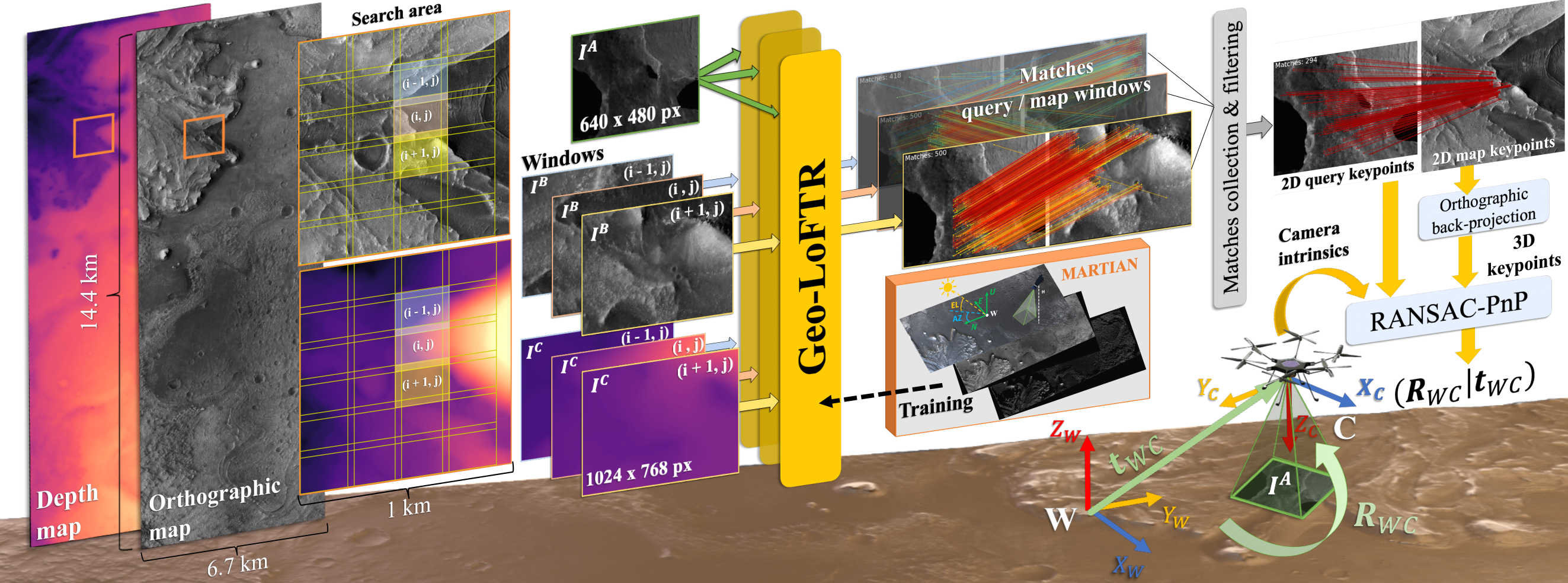}
    \caption{Given orthographic grayscale and depth maps geo-referenced in the global frame $W$, and an onboard image $I^A$ from the navigation camera $C$ of a Mars rotorcraft, we aim to estimate the pose ($\mathbf{R}_{WC} \vert \mathbf{t}_{WC}$). From a noisy pose prior, a search area is defined and divided into map $I^B(i,j)$ and depth $I^C(i,j)$ windows. These form triplets $I^A, I^B(i,j), I^C(i,j)$ that are sequentially processed by our geometry-enhanced Geo-LoFTR matcher, trained on data from our MARTIAN simulation tool. The resulting matches are filtered and used in RANSAC-PnP to recover the rotorcraft pose.
    }
    \vspace{-1pt}
    \label{fig:CONCEPT}
\end{figure*}

In this paper, we explore a new MbL system that makes no assumptions on illumination conditions or
scale variation for vision-based geo-localization on Mars for future helicopters. In particular, we incorporate a transformer-based method~\cite{loftr} for image registration into the MbL pipeline, and enhance it by leveraging terrain geometry cues to increase robustness to lighting variations. Furthermore, we developed the Mars Aerial Rendering Tool for Imaging and Navigation (MARTIAN) to generate a large-scale Mars dataset from orbital maps and train the image registration method. In summary, our contributions towards a robust MbL pipeline include:
\begin{itemize}
    \item A new image matching method, Geo-LoFTR, using geometric context from Digital Terrain Models (DTMs) and improving localization accuracy over prior methods @ 1m up to 31.8\% in challenging illumination conditions. 
    \item A custom simulation pipeline to generate maps and aerial observations of realistic Martian landscapes derived from real orbital data products.  
    \item A comprehensive evaluation of our MbL approach on synthetic and real Mars imagery. 
\end{itemize}

\section{RELATED WORK}
\label{sec:related_work}

Space-based applications of MbL traditionally rely on template-matching techniques, where an ortho-rectified onboard image slides over a reference map to estimate pixel-wise similarity using distance metrics such as Normalized Cross-Correlation~\cite{pham2021rover}, Phase-Correlation~\cite{WAN2016198}, and Mutual Information~\cite{ansar2009multi}.
The Mars2020 Lander Vision System (LVS) employed a coarse-to-fine template-matching approach to perform absolute localization on a Context Camera (CTX) map (6 m/px) during the Entry Descent and Landing (EDL) phase \cite{johnson2023}.
The Censible framework proposed in \cite{nash2024} successfully performed global localization onboard the Perseverance rover by registering an ortho-mosaic of panoramic stereo images to a HiRISE map (0.25 m/px) with a modified census transform~\cite{zabih1994non}. While effective in the specific cases, these methods operated under minimal lighting variations and are not robust to viewpoint changes and in-plane rotation without a correction step. Recently, the Lighting Invariant Matching Algorithm~\cite{rothenberger2025illumination} proposed a correlation-based method robust to challenging illumination, but it assumes at least two pre-registered images with lighting diversity.

Beyond template-matching, hand-crafted features such as SIFT~\cite{sift},~ORB \cite{orb}, SURF~\cite{surf}, and SOSE~\cite{cheng2024simultaneous} have been investigated and pitted against deep learning approaches for MbL and related problems. For example, the work of~\cite{brockers2022} demonstrated SuperPoint~\cite{superpoint} to outperform SIFT in localization accuracy at very low sun elevations, while AstroVision~\cite{driver2023astrovision} showed
ASLFeat~\cite{luo2020aslfeat} to be more robust than hand-crafted features for feature tracking under large shadows for visual navigation around small bodies. JointLoc~\cite{luo2024jointloc} proposed a Mars UAV localization system using SuperPoint~\cite{superpoint} and LightGlue~\cite{lindenberger2023lightglue} for local feature matching, but without testing challenging illumination or scale variations, and relying mainly on synthetic data.

Many deep learning methods for image matching have shown robustness to real-world changes in scale, illumination, and viewpoint. LoFTR~\cite{loftr} uses Transformers~\cite{transformer} in a detector-free manner, RoMa~\cite{roma} leverages features from the foundation model of DINOv2~\cite{dinov2}, and DKM~\cite{edstedt2023dkm} estimates dense warps for pixel-wise correspondences. Other works such as GAM~\cite{yu2022improving} and GoMatch~\cite{zhou2022geometry} sought to introduce geometric information in image matching for localization tasks. Inspired by these recent developments, we adapt the state-of-the-art method of LoFTR~\cite{loftr} for MbL on Mars and extend its architecture to leverage geometric context. Finally, we acknowledge the 
body of work on vision-based global localization for UAVs in Earth-based applications~\cite{couturier2024review, Xu2018VisionbasedUA}. 
\section{METHODOLOGY}
\label{sec:methodology}
\begin{figure}
    \centering
    \vspace{3pt}
    \includegraphics[width=\linewidth]{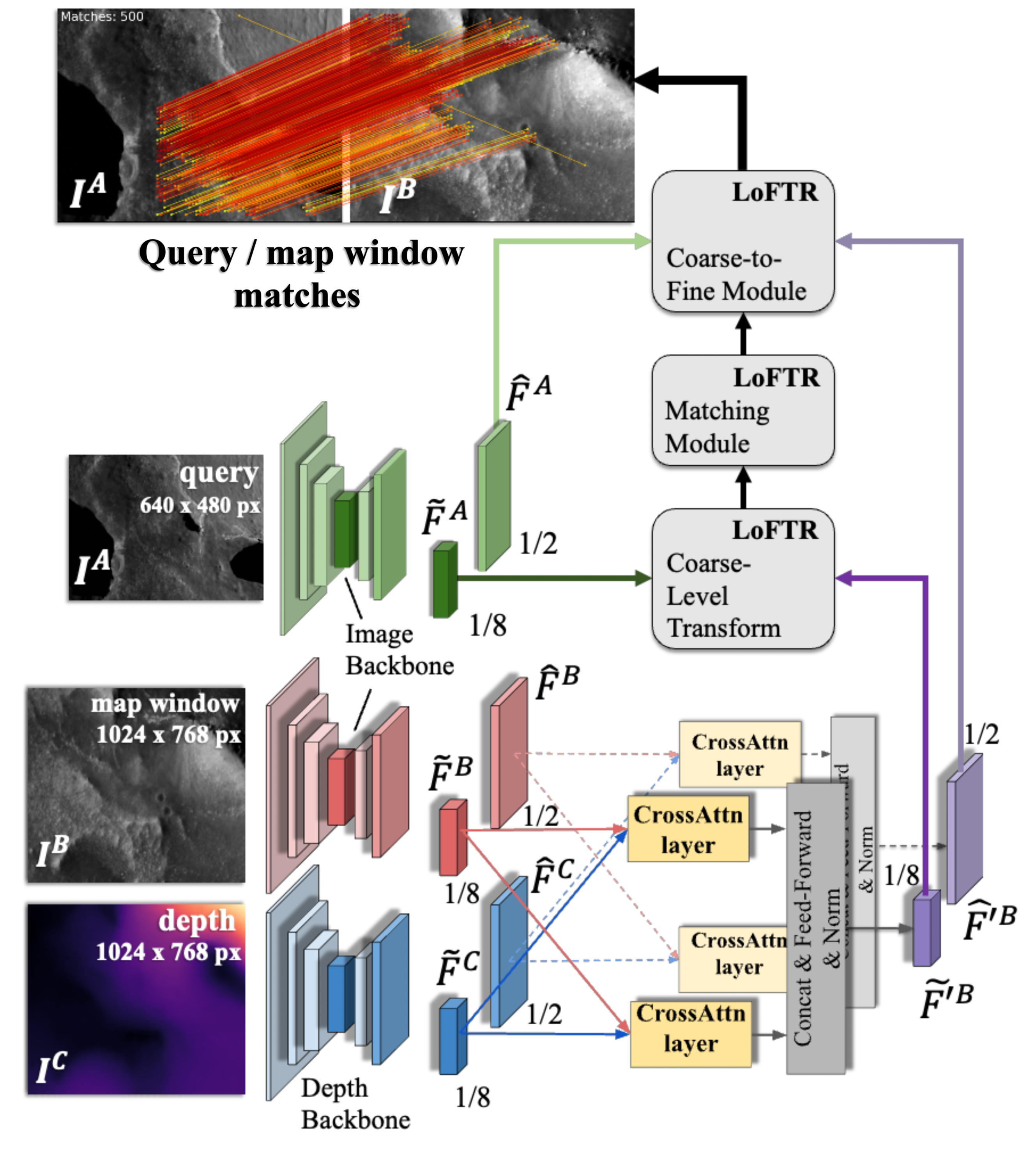}
    \vspace{-16pt}
    \caption{Architecture of Geo-LoFTR that uses as inputs the query image $I^A$ and a map crop image $I^B$ along with its corresponding depth $I^C$. Geo-LoFTR learns to merge visual information from $I^B$ with geometric information from $I^C$ using parallel $CrossAttn$ operations. The produced features then follow the coarse-to-fine approach proposed in LoFTR~\cite{loftr}. Our experimental evaluation shows that Geo-LoFTR is more robust than the original LoFTR under challenging illumination conditions.}
    \label{fig:geo-loftr}
\end{figure}
We propose a new MbL method that is robust to challenging illumination conditions on Mars.
We introduce Geo-LoFTR, an image matching model that learns to merge geometric and visual features (Sec.~\ref{subsec:geo-loftr}). 

The base method, LoFTR~\cite{loftr}, can offer robustness to images with large shadows by using transformers that draw information from other parts of the image. Moreover, its use of linear attention layers \cite{katharopoulos2020transformers} makes it more computationally efficient than other transformer-based approaches (e.g., RoMa~\cite{roma}). On our Mars dataset, the LoFTR model pre-trained on the MegaDepth dataset also compared favorably with other popular methods such as SuperGlue~\cite{sarlin2020superglue} and LightGlue~\cite{lindenberger2023lightglue}, balancing computational complexity and localization accuracy across a wide variety of illumination conditions (Fig.~\ref{fig:cdf_sun_el_var}, ~\ref{fig:cdf_sun_az_var}).
These properties make LoFTR a suitable base model for our geometry-aided image registration. 

We also present MARTIAN, our aerial simulation tool, (Sec.~\ref{subsec:martian}), and outline our dataset generation strategy (Sec.~\ref{subsec:training_set}). Finally, we discuss the MbL pipeline (Sec.~\ref{subsec:mbl_pipeline}).
An overview of our framework is illustrated in Fig.~\ref{fig:CONCEPT}.

\subsection{\label{subsec:geo-loftr}Geometry-aided Local Feature Matching}
\noindent \textbf{Preliminaries.}
LoFTR follows a detector-free approach to produce semi-dense matches between two images $I^A$ and $I^B$. Its strength lies on using transformers to process features from a CNN backbone at two scales $\tilde{F}^A \in \mathbb{R}^{\frac{H}{8} \times \frac{W}{8} \times d}$ and $\hat{F}^A \in \mathbb{R}^{\frac{H}{2} \times \frac{W}{2} \times d}$, where $H,W$ are height and width of $I^A$ and $d$ is the feature dimension. This allows the semantically rich pixel-wise representations to capture global image context. The corresponding features are also extracted from $I^B$. The method then follows a coarse-to-fine approach, where initial dense matching is performed at the coarse level, followed by refinement of the matches around a small window using the higher resolution feature maps. For more details, see the original paper~\cite{loftr}.

\noindent \textbf{Geo-LoFTR.}
We aim to incorporate geometric context during feature matching between onboard observations and the reference map to increase robustness to challenging lighting scenarios where visual cues alone may lead to degeneracy.
In the context of MbL on Mars, we take advantage of the 3D information from a DTM and enrich the learned representation of the ortho-projected map.

To accomplish this, we extend the original LoFTR architecture to take as input a depth image $I^C$ of a map crop along with the corresponding grayscale image $I^B$ of the crop from the ortho-rectified map and the grayscale onboard image $I^A$. Each input depth map crop is normalized using the highest depth value in the crop, to avoid overfitting on absolute depth values of local areas, but rather associate relative geometry with visual information. 
In order to extract the corresponding coarse $\tilde{F}^C$ and fine $\hat{F}^C$ features for $I^C$ we use the same ResNet-18~\cite{resnet-18} backbone from LoFTR.

Our objective is to learn how to merge the features $\tilde{F}^B$, $\hat{F}^B$ with the corresponding $\tilde{F}^C$, $\hat{F}^C$ to produce coarse- and fine-level feature maps that incorporate both visual and geometric information of the local map area.
To do so, we use cross-attention layers in both directions:
\begin{align*}
     \tilde{F}'^B = G \left(CrossAttn(\tilde{F}^B, \tilde{F}^C) \oplus CrossAttn(\tilde{F}^C, \tilde{F}^B) \right)
\end{align*}
where the first argument for each $CrossAttn$ layer is used as the query, $\oplus$ concatenates the outputs along the feature dimension, and $G$ is a small feedforward network comprised of two linear layers followed by LayerNorm.
The resulting $\tilde{F}'^B \in \mathbb{R}^{\frac{H}{8} \times \frac{W}{8} \times d}$ is the merged coarse level feature map. The corresponding process is repeated for the fine level maps to produce $\hat{F}'^B \in \mathbb{R}^{\frac{H}{2} \times \frac{W}{2} \times d}$. The rest of the pipeline follows the original LoFTR, with the merged feature representations being used in the coarse and fine transformer modules instead of the features extracted only from $I^B$. 
Fig. \ref{fig:geo-loftr} shows the architectural details of Geo-LoFTR.

\subsection{MARTIAN: Mars Aerial Rendering Tool for Imaging and Navigation}
\label{subsec:martian}
Unlike Earth-based applications that have the privilege of abundant data~\cite{zhu2023sues}, relevant, annotated, and large-scale aerial datasets are not readily available for Mars. Data released from the Mars2020 mission offer aerial observations from the LVS camera during EDL and from the navigation camera of Ingenuity. However, these data lack accurate pose annotations and are insufficient for a comprehensive robustness study on scale and illumination variations.

Instead, we leverage real map products derived from the Mars Reconnaissance Orbiter (MRO) High-Resolution Imaging Science Experiment (HiRISE)~\cite{hirise} to create a large-scale dataset for training Geo-LoFTR and evaluating our MbL pipeline. We developed a Python-based framework in Blender that imports HiRISE products to generate maps and aerial observations of a Martian site under varying lighting conditions and altitudes.
In this work, we use a 1 m/post DTM and a 0.25 m/px ortho-image as high-fidelity texture, generated from stereo imaging of the Jezero crater site over an area of 7 km by 14 km. The generated mesh is assigned a material with surface shading defined by a Principled Bidirectional Scattering Distribution Function (BSDF), tuned to match the visual appearance of the real Jezero map to MARTIAN renderings under identical illumination.

\begin{figure}
    \centering
    \vspace{3pt}
    \includegraphics[width=\linewidth]{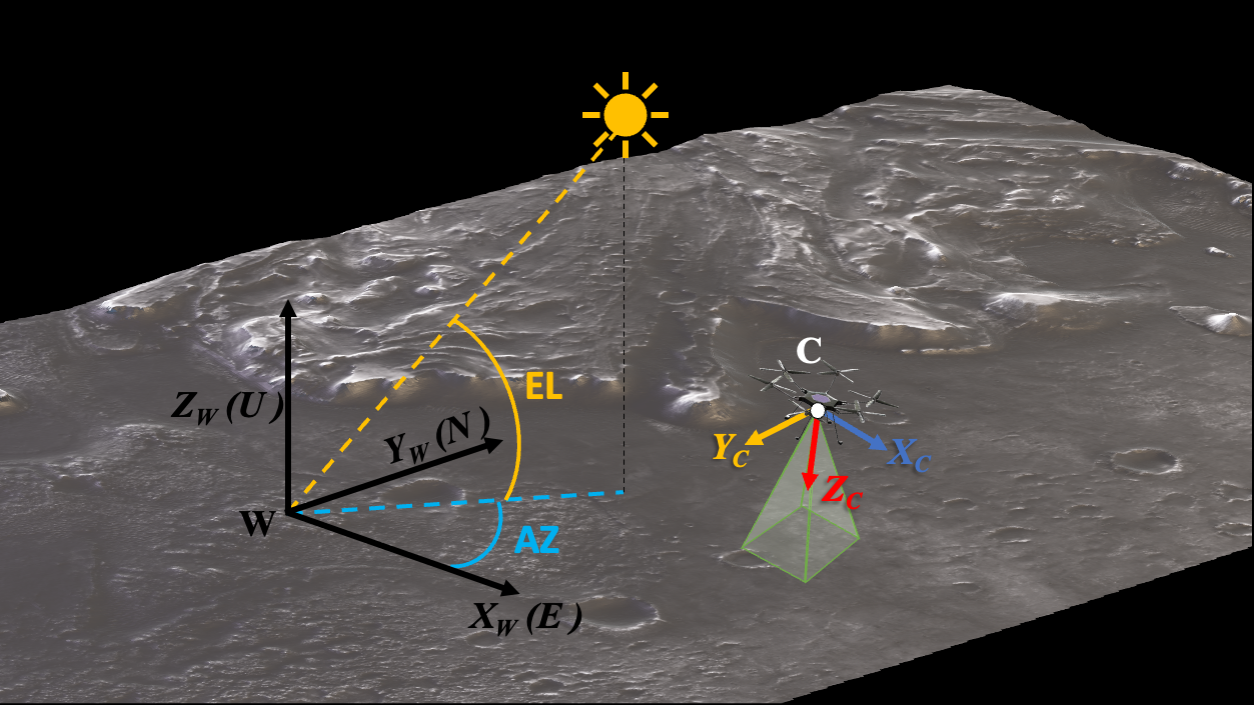}
    \vspace{-16pt}
    \caption{\label{fig:MARTIAN_view}View of the Jezero Crater's DTM in MARTIAN. A \textit{world} reference frame, $W$, is defined as a East-North-Up coordinate system with origin on the map center. A \text{camera} frame $C$ is centered at the camera's optical center, with its X-axis pointing to the right along the image width, and the Z-axis pointing towards the terrain along the camera optical axis.}
\end{figure}

Scenes can be rendered in perspective or orthographic mode with user-defined camera intrinsics and extrinsics. The camera pose is defined as ($\mathbf{R}_{WC} \vert \mathbf{t}_{WC}$), where $\mathbf{t}_{WC}$ is the location of the camera frame, $C$, in world coordinates, $W$, and $\mathbf{R}_{WC}$ is the rotation matrix that aligns $W$ to $C$ (Fig.~\ref{fig:MARTIAN_view}). The time of day is simulated by specifying the Sun elevation (EL) and azimuth angles (AZ), while brightness and shadow softness are controlled by Blender solar irradiance and angular disk parameters, consistent with Mars estimates. In concurrent work~\cite{georgakis2025}, LoFTR models trained on MARTIAN successfully registered Ingenuity navigation imagery to HiRISE maps, supporting the realism of renderings derived from orbital products.

\subsection{Generating a training set with MARTIAN}
\label{subsec:training_set}
We generated a training image dataset with 17 orthographic grayscale maps rendered with sun azimuth spanning $0^{\circ}$-$360^{\circ}$ and elevations of $30^{\circ}$, $60^{\circ}$, $90^{\circ}$. Lower elevations down to $5^{\circ}$ are reserved for out-of-distribution evaluation in Sec. \ref{sec:mbl_eval}. We also generated one corresponding orthographic depth map and 4500 nadir aerial observations at fixed sun AZ=180$^{\circ}$, EL=40$^{\circ}$ with depth images. The query observations were randomly sampled from the HiRISE Jezero crater DTM in the 64-200 m altitude range, covering MSH nominal operations. The 64 m minimum reflects MARTIAN's resolution limit, which matches the HiRISE texture.

Geo-LoFTR training uses triples ($I_A$, $I_B$, $I_C$) comprising a query ($I_A$), and co-located grayscale ($I_B$) and depth ($I_C$) map windows crops at multiple solar angle offsets between queries and source maps. 
For each  offset, each query is paired with multiple map windows with at least 25\% overlap with the query footprint.
Window sizes are chosen to induce an appropriate scale variation across the altitude range while balancing generalization and training efficiency. 
Geo-LoFTR is fine-tuned from the original LoFTR pre-trained model on 150K generated triplets. A validation set of 3400 triplets monitors performance and prevents overfitting.

\subsection{Map-based Localization Pipeline}
\label{subsec:mbl_pipeline}
The goal of map-based localization is to retrieve the onboard (query) camera pose $(\mathbf{R}_{WC_{\text{query}}}, \vert \mathbf{t}_{WC_{\text{query}}})$ in globalframe $W$ (Fig.~\ref{fig:MARTIAN_view}).
This is preceded by the registration of the query observation captured by a camera with known intrinsic parameters over a geo-referenced ortho-projected map.

We assume that future Mars rotocraft will be equipped with onboard VIO such that a noisy pose prior would be provided to the MbL pipeline. This narrows the registration to a search area sized by the VIO pose uncertainty. In our work, we assume a large predetermined search area of 1 km$^2$ centered at the query observation for two reasons. First, it simulates a conservative scenario with a high-uncertainty pose prior. Second, it decouples localization performance from pose prior quality across experiments.


For each query, the map search area is further divided into multiple windows of size $1024 \times 768$ pixels and with 10\% overlap. The query image $I^A$ is paired with each map window crop $I^B$ and the corresponding depth image crop $I^C$. The formed triplet $(I^A, I^B, I^C)$ is processed by Geo-LoFTR, which outputs matched keypoints on both map windows and queries with their confidence scores. The top 500 matches per window are further filtered to a 95\% confidence threshold across the search area.
Given a simulated orthographic map camera with pose $(\mathbf{R}_{WC_{\text{map}}}, \vert \mathbf{t}_{WC_{\text{map}}})$ in the world frame, each matched  point of pixel coordinates $(u_{\text{map}}, v_{\text{map}})$ in the map image plane is back-projected to its 3D location in $W$ using an inverse orthographic projection:
\begin{equation*}
     \begin{bmatrix} X^W \\ Y^W \\ Z^W \end{bmatrix} = p_{\text{map}} \mathbf{R}_{WC_{\text{map}}} \, \begin{bmatrix} 
                    1 & 0 & -c_{x,{\text{map}}} \\
                    0 & 1 &  - c_{y,{\text{map}}}  \\
                    0 & 0 &Z/p_{\text{map}}  \end{bmatrix} \begin{bmatrix} u_{\text{map}} \\ v_{\text{map}} \\ 1
                    \end{bmatrix} + \mathbf{t}_{WC_{\text{map}}}
    \label{eq:backproj_map_pts}
\end{equation*}
where $Z$ is the depth of the map keypoint,  $(c_{x,{\text{map}}}, c_{y,{\text{map}}})$ is the optical center and $p_{\text{map}}$ is the pixel resolution. 
The matched 2D-3D correspondences and camera intrinsics $\mathbf{K}$ are fed to a RANSAC-PnP solver to
retrieve the estimated pose $(\widetilde{\mathbf{R}}_{WC_{\text{query}}}, \vert \widetilde{\mathbf{t}}_{WC_{\text{query}}})$ via perspective projection.
\section{EXPERIMENTAL EVALUATION}
\label{sec:mbl_eval}

We evaluated our MbL pipeline on MARTIAN-generated datasets from HiRISE maps of the Jezero crater and real Mars2020 descent imagery, ensuring no overlap between training and test data for an unbiased assessment.

We compared our results with LoFTR fine-tuned on our dataset (\textit{Fine-LoFTR}), LoFTR pre-trained on MegaDepth (\textit{Pre-LoFTR}), and SIFT, one of the most accurate handcrafted methods for localization on simulated Mars terrain~\cite{brockers2022}.
We also evaluated SuperGlue (\textit{Pre-SP+SG})~\cite{sarlin2020superglue} and LightGlue~\cite{lindenberger2023lightglue} (\textit{Pre-SP+LG}) with SuperPoint~\cite{superpoint} features pre-trained on MegaDepth to justify the choice of our base-model for the geometric extension. We focus on comparing feature matching approaches rather than visual place recognition methods, which assume densely sampled database views of the terrain in the same domain as the query. In contrast, our task involves cross-view orbital-to-UAV matching using lower resolution maps.

For Ingenuity, the desired accuracy was set to 5m~\cite{anderson2024lessons}, though this might evolve in future missions. We therefore report the Cumulative Distribution Function (CDF) of the localization error up to 10 m.

\begin{figure}
\centering
\vspace{3pt}
\begin{minipage}[b]{0.49\linewidth}
    \centering
    \includegraphics[width=\linewidth]{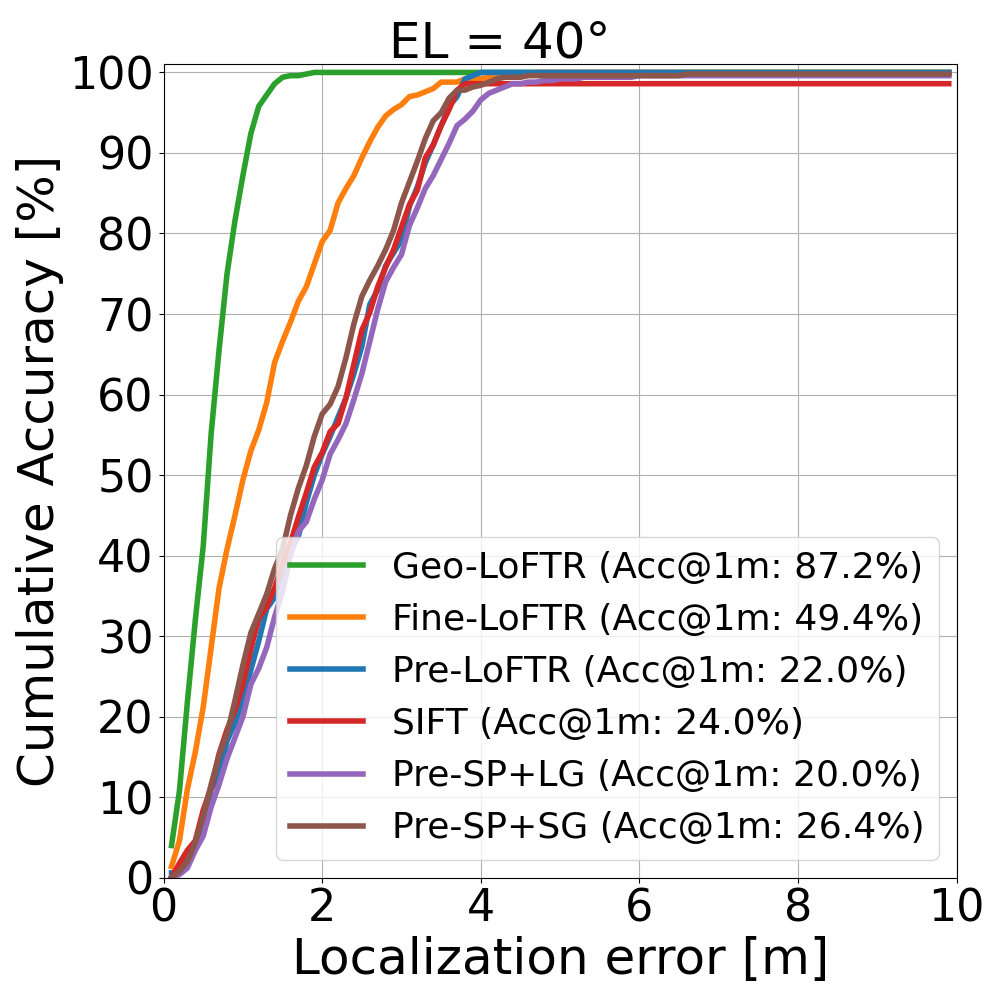}
\end{minipage}
\begin{minipage}[b]{0.49\linewidth}
    \centering
    \includegraphics[width=\linewidth]{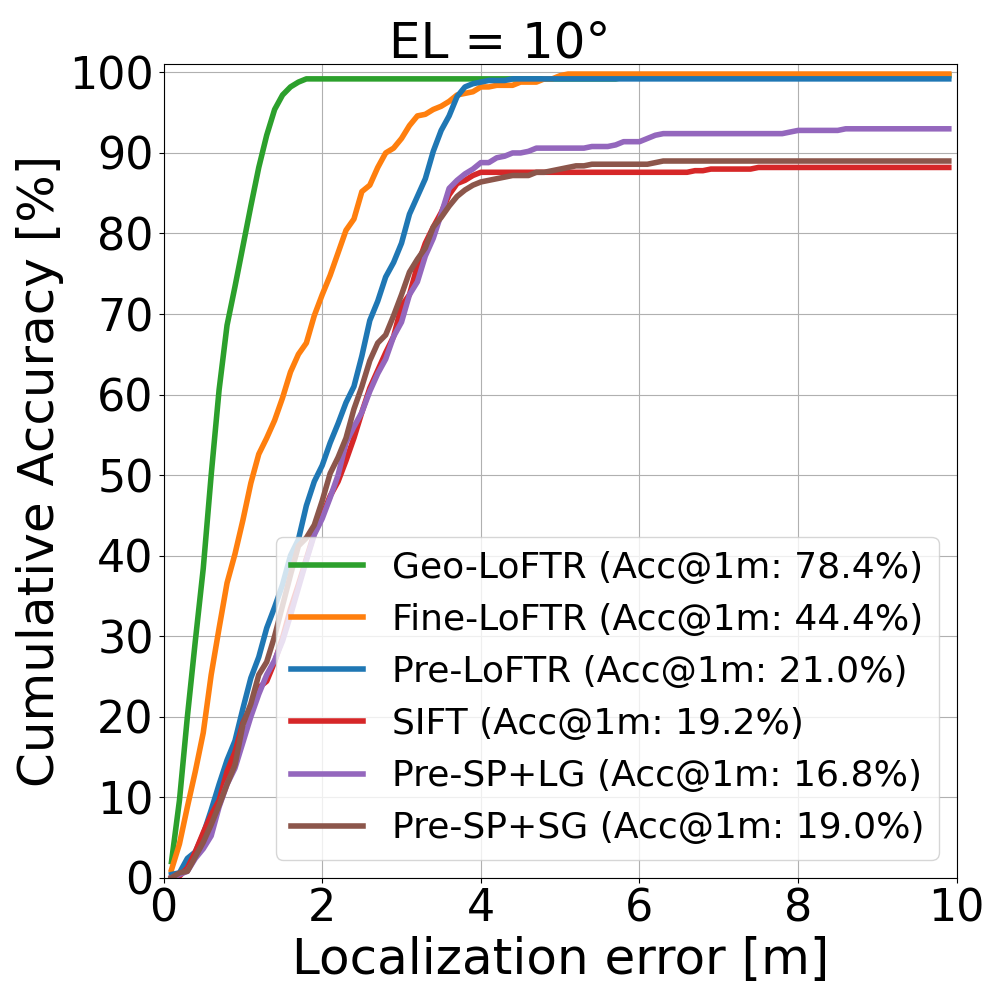}
\end{minipage}
\\
\begin{minipage}[b]{0.49\linewidth}
    \centering
    \includegraphics[width=\linewidth]{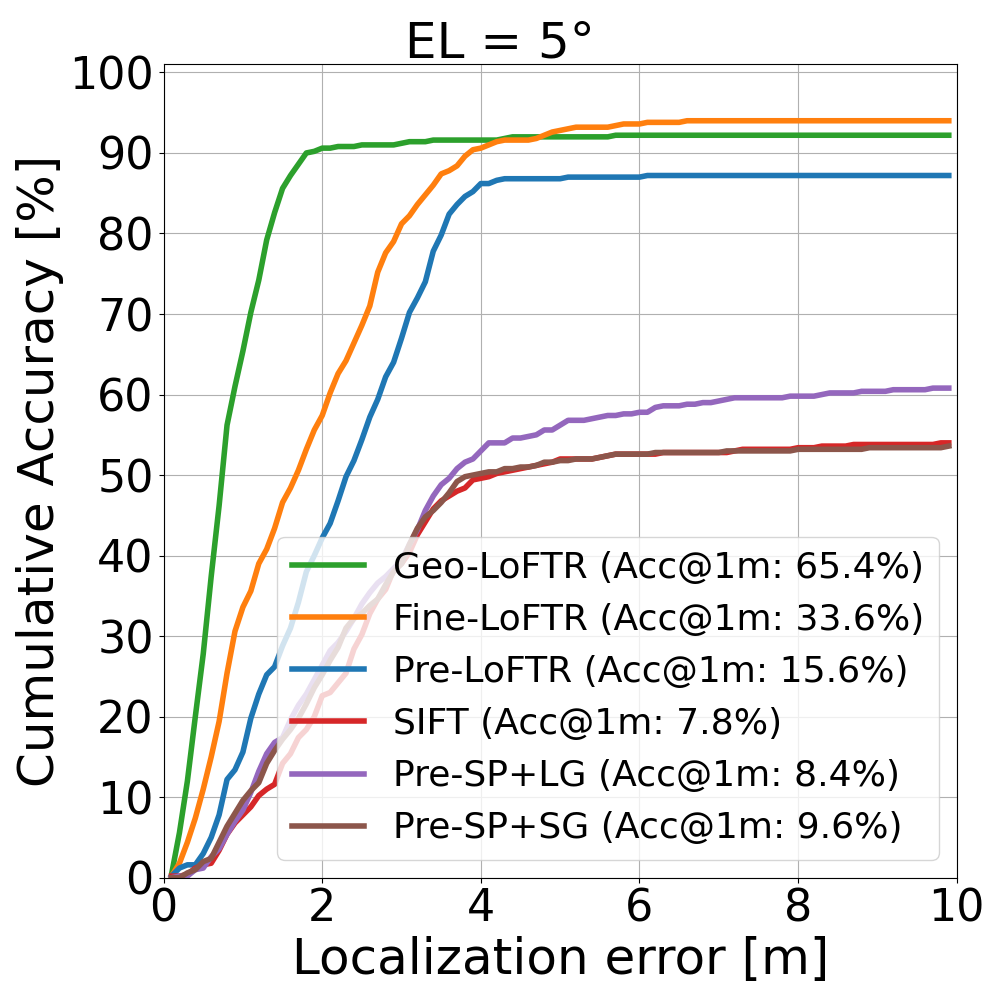}
\end{minipage}
\begin{minipage}[b]{0.49\linewidth}
    \centering
    \includegraphics[width=\linewidth]{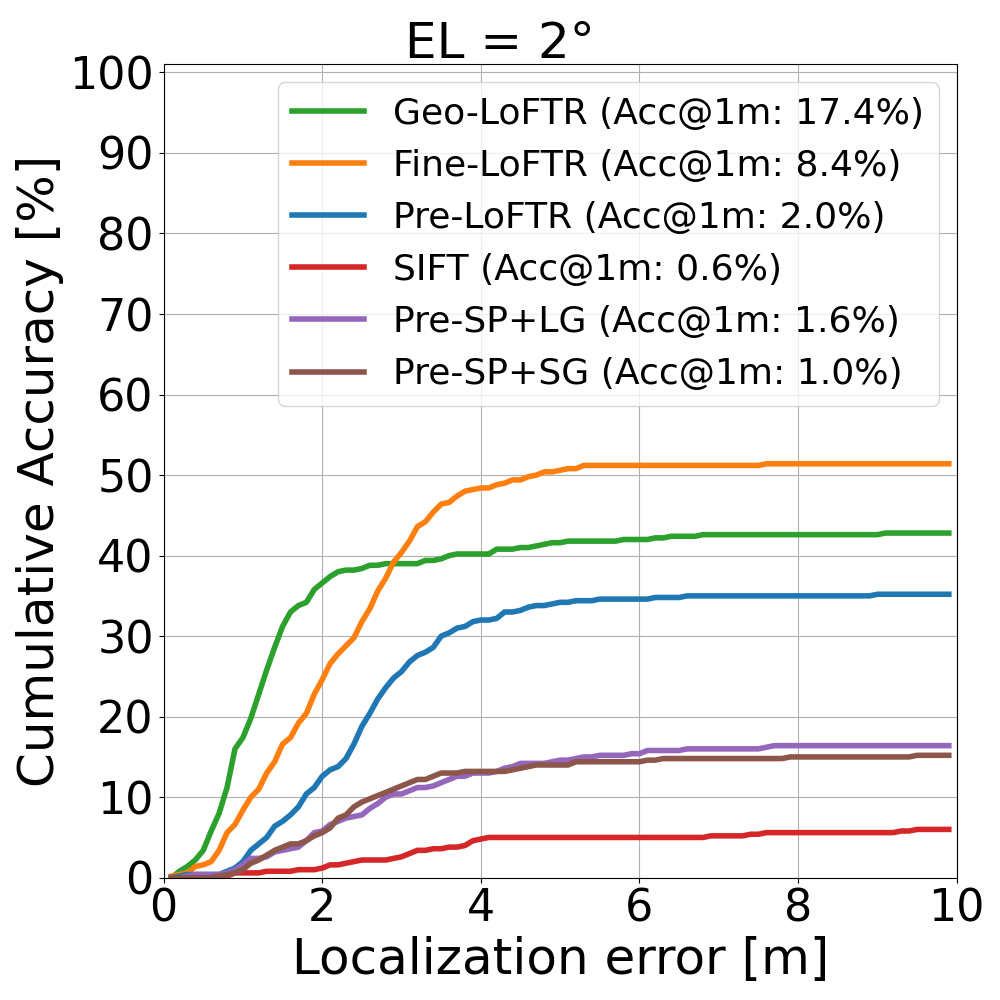}
\end{minipage}
\vspace{-16pt}
\caption{\label{fig:cdf_sun_el_var}Cumulative distributions of the localization error of simulated Mars observations in the 64-200 m altitude range at sun AZ=180$^{\circ}$ and EL=40$^{\circ}$, registered onto maps at four different elevation angles and 0$^{\circ}$ azimuth offset.}
\end{figure}
\subsection{Robustness to Changing Solar Angles}
\label{subsec:mbl_robustnbess_to_az_and_el}
We registered query observations onto orthographic maps rendered with varying sun elevation and azimuth angles, evaluating each effect separately. Although this setup differs from the operational scenario (Sec. \ref{subsec:mbl_martian_day}), varying illumination on the map side introduces controlled sun AZ and EL offsets between maps and observations, while enabling efficient dataset generation across multiple lighting conditions. Finally, we assess robustness to scale variation.
\\

\noindent \textbf{Robustness to sun elevation.}
For this experiment, our map test set comprises orthographic images rendered at EL=$\{2°, 5°, 10°, 40°, 60°, 90°\}$ and AZ=$180°$, along with 500 nadir-pointing queries at fixed sun angles (AZ=$180°$, EL=$40°$), and altitudes between 64 and 200 m.
We note that elevations below 30$^{\circ}$ were not encountered during model training. On the day of our reference HiRISE acquisition, the Sun elevation spanned 7.9-82.6$^{\circ}$, from 6:00 to 17:00 Local Mean Solar Time (LMST).
Elevations $<10^{\circ}$ cover only a brief portion of the local solar trajectory, representing exceptional cases for Martian surface operations. Nevertheless, we include them to assess generalization under challenging lighting conditions. 

Fig.~\ref{fig:cdf_sun_el_var} shows the CDFs of the localization error for four elevation offsets at the same azimuth. Geo-LoFTR outperforms the baselines, indicating that incorporating depth data mitigates degeneracies in purely visual data.
With no lighting difference between maps and observations, Geo-LoFTR is 87\% @1m accurate, showing an improvement of +38\% over the fine-tuned model, and +63\% over SIFT. Below 10$^{\circ}$ EL, all the methods significantly degrade. While Geo-LoFTR remains the most accurate at the very extreme case of 2$^{\circ} $EL, poor illumination and extensive shadow coverage may saturate the geometric information.
\\

\begin{figure}
\centering
\vspace{3pt}
\begin{minipage}[b]{0.49\linewidth}
    \centering
    \includegraphics[width=\linewidth]{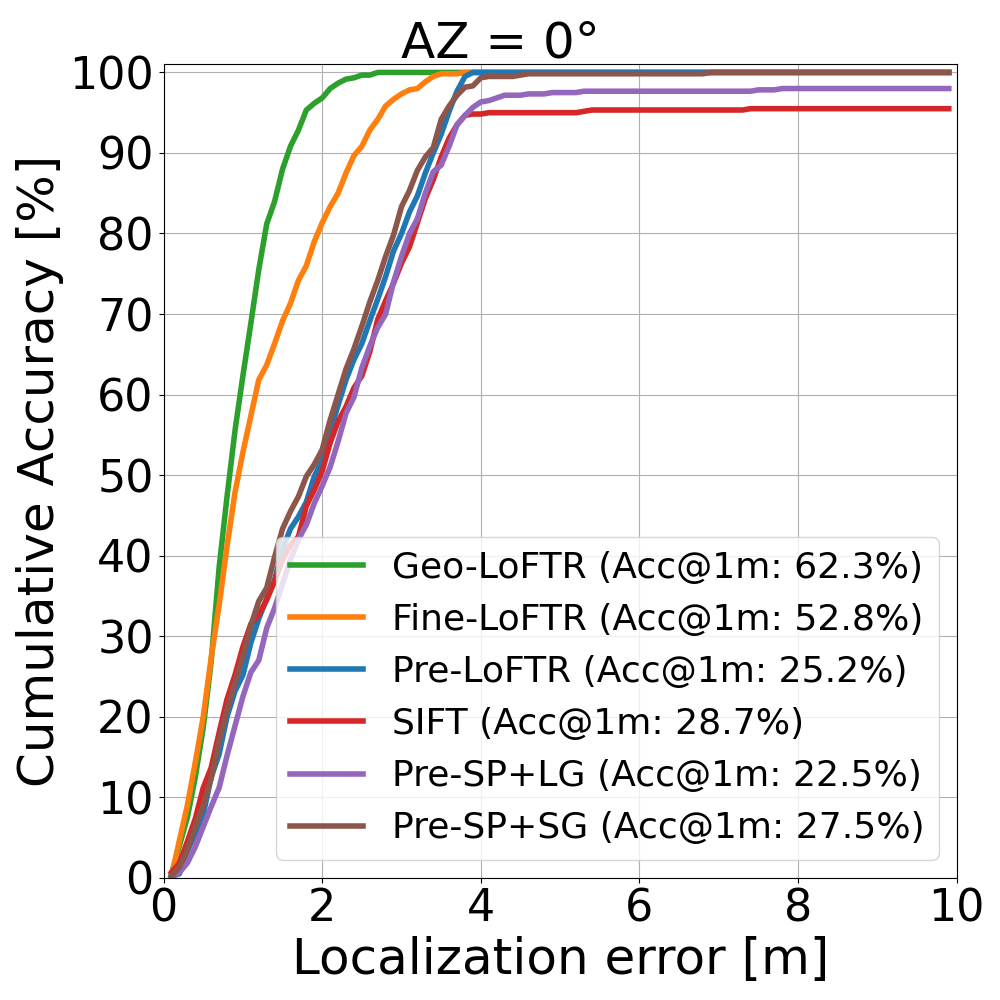}
\end{minipage}
\begin{minipage}[b]{0.49\linewidth}
    \centering
    \includegraphics[width=\linewidth]{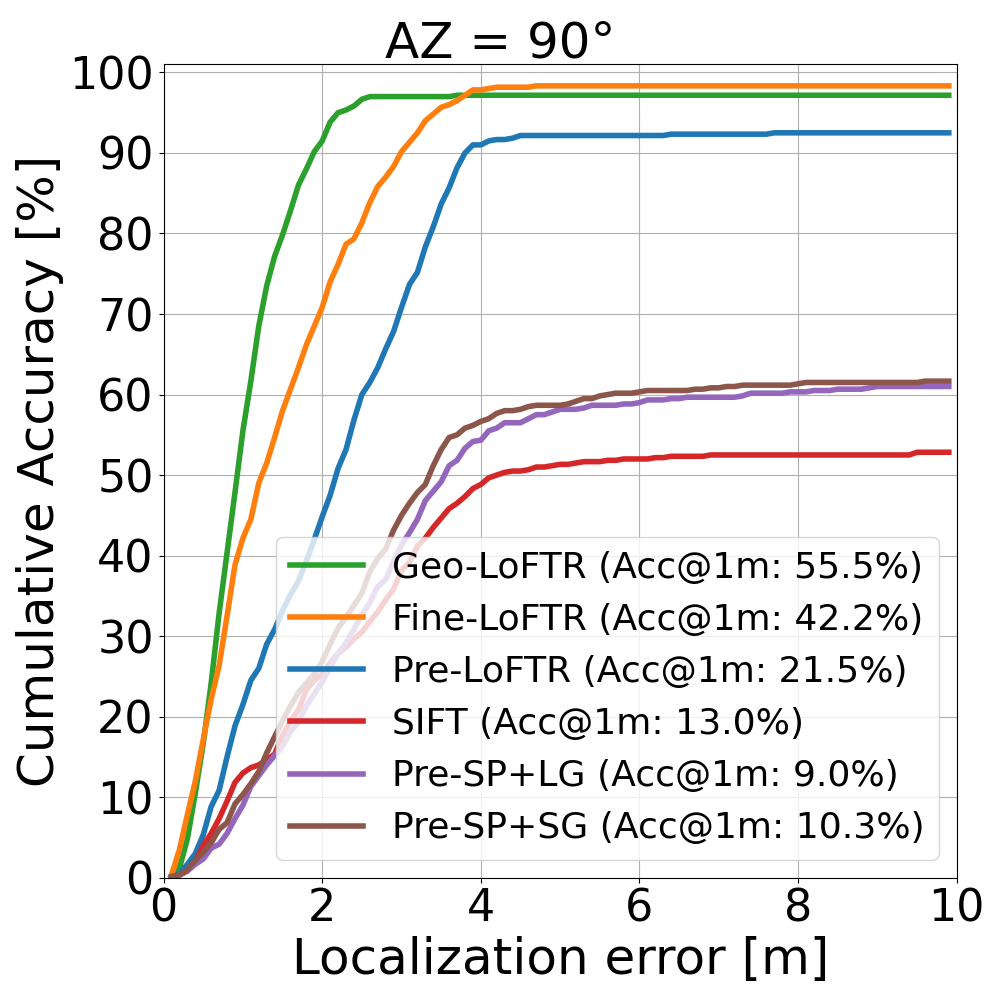}
\end{minipage}
\\\vspace{2pt}
\begin{minipage}[b]{0.49\linewidth}
    \centering
    \includegraphics[width=\linewidth]{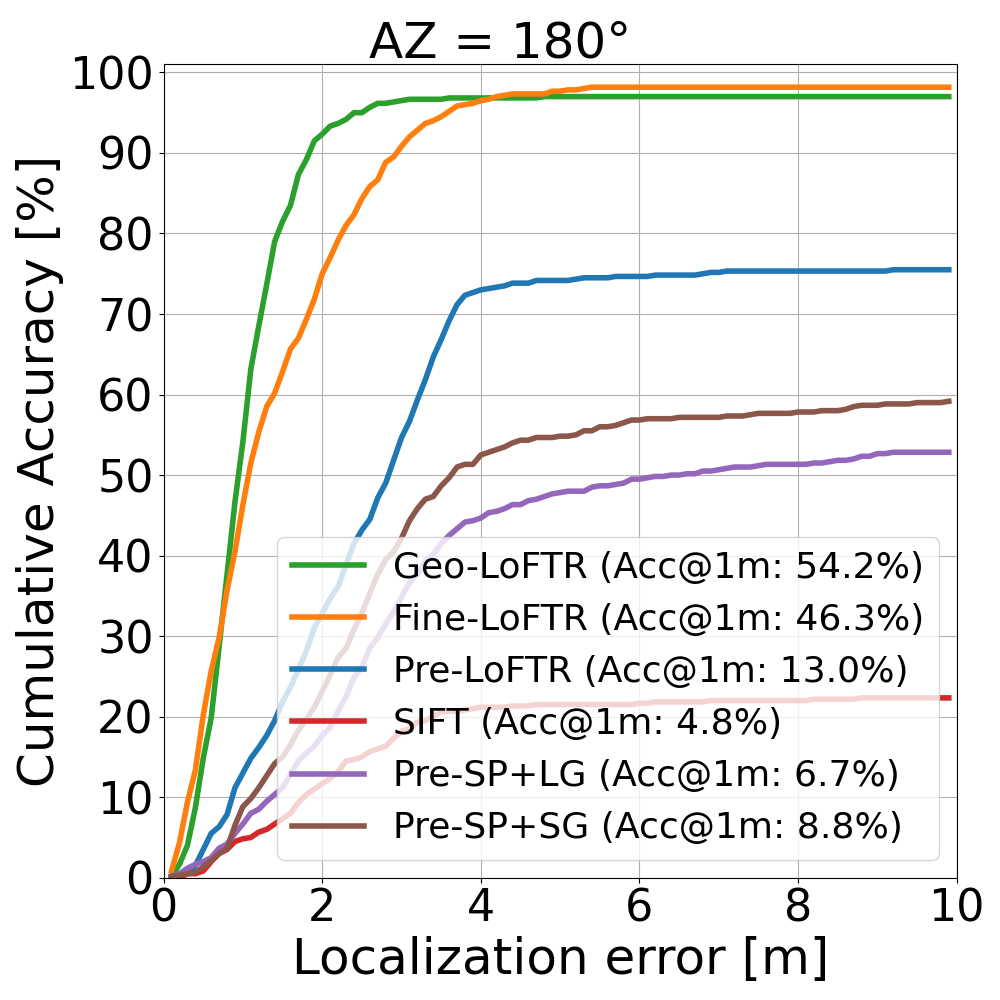}
\end{minipage}
\begin{minipage}[b]{0.49\linewidth}
    \centering
    \includegraphics[width=\linewidth]{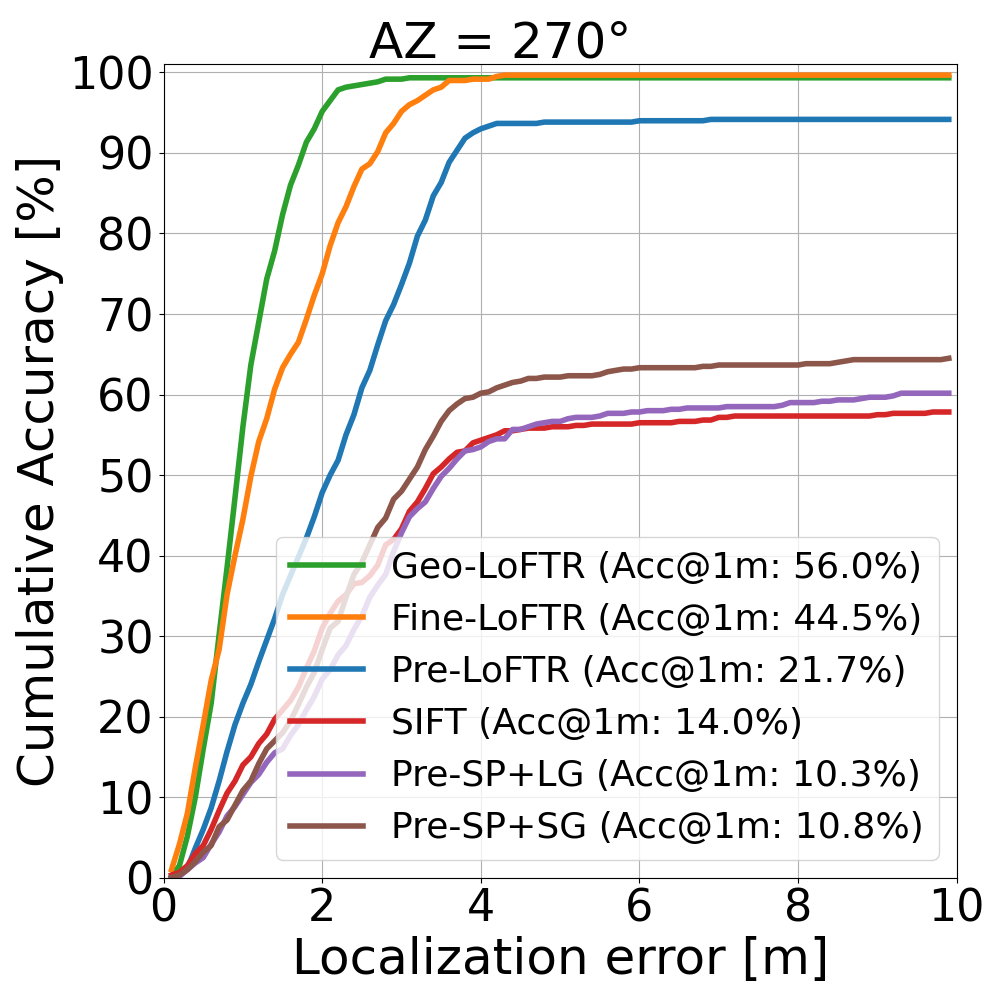}
\end{minipage}
\vspace{-16pt}
\caption{\label{fig:cdf_sun_az_var}Cumulative distributions of the localization error of simulated Mars observations in the 64-200 m altitude range at sun AZ=0$^{\circ}$ and EL=10$^{\circ}$, registered onto maps at four different azimuth angles and 0$^{\circ}$ elevation offset.}
\end{figure}

\begin{figure*}
\vspace{3pt}
\centering
\includegraphics[width=0.85\linewidth]{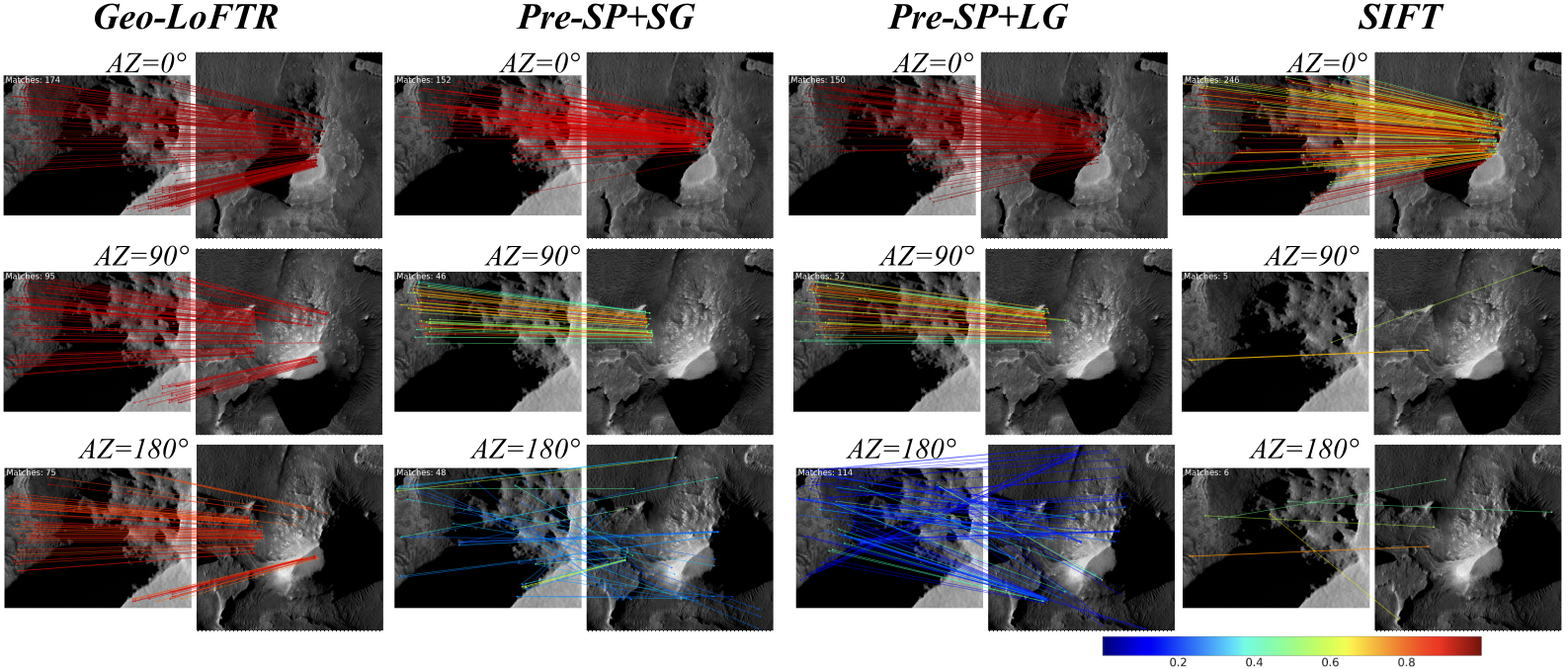}
\vspace{-7pt}
\caption{Geo-LoFTR, Pre-SP+SG, Pre-SP+LG and SIFT matched keypoints displayed for a sample query image (\textit{left side of each panel}) with (0$^{\circ}$ AZ, 10$^{\circ}$ EL) sun angles and a map search area image (\textit{right side of each panel}) under three different sun azimuths and 0$^{\circ}$ elevation offset. Match lines are color-coded by confidence score, with redder indicating higher confidence.
}
\label{fig:matches_vs_az}
\end{figure*}

\noindent \textbf{Robustness to sun azimuth.}
We generated 500 nadir-pointing queries (AZ=0$^{\circ}$, EL=10$^{\circ}$) to be registered onto maps with varying azimuth angles across 0-360$^{\circ}$ and same elevation. 
Fig.~\ref{fig:cdf_sun_az_var} shows the CDFs  of the localization error for four different sun azimuth offsets. Geo-LoFTR proved to be the most accurate model with a @1m accuracy in the 54-63 \% range, despite the relatively low sun elevation of 10$^{\circ}$.
The number and quality of the SIFT matched keypoints between query and map (Fig.~\ref{fig:matches_vs_az}) decrease faster than LoFTR-based models as we depart from zero AZ offset.
\\


\begin{figure}
\centering
\begin{minipage}[b]{1.0\linewidth}
    \centering
    \includegraphics[width=\linewidth]{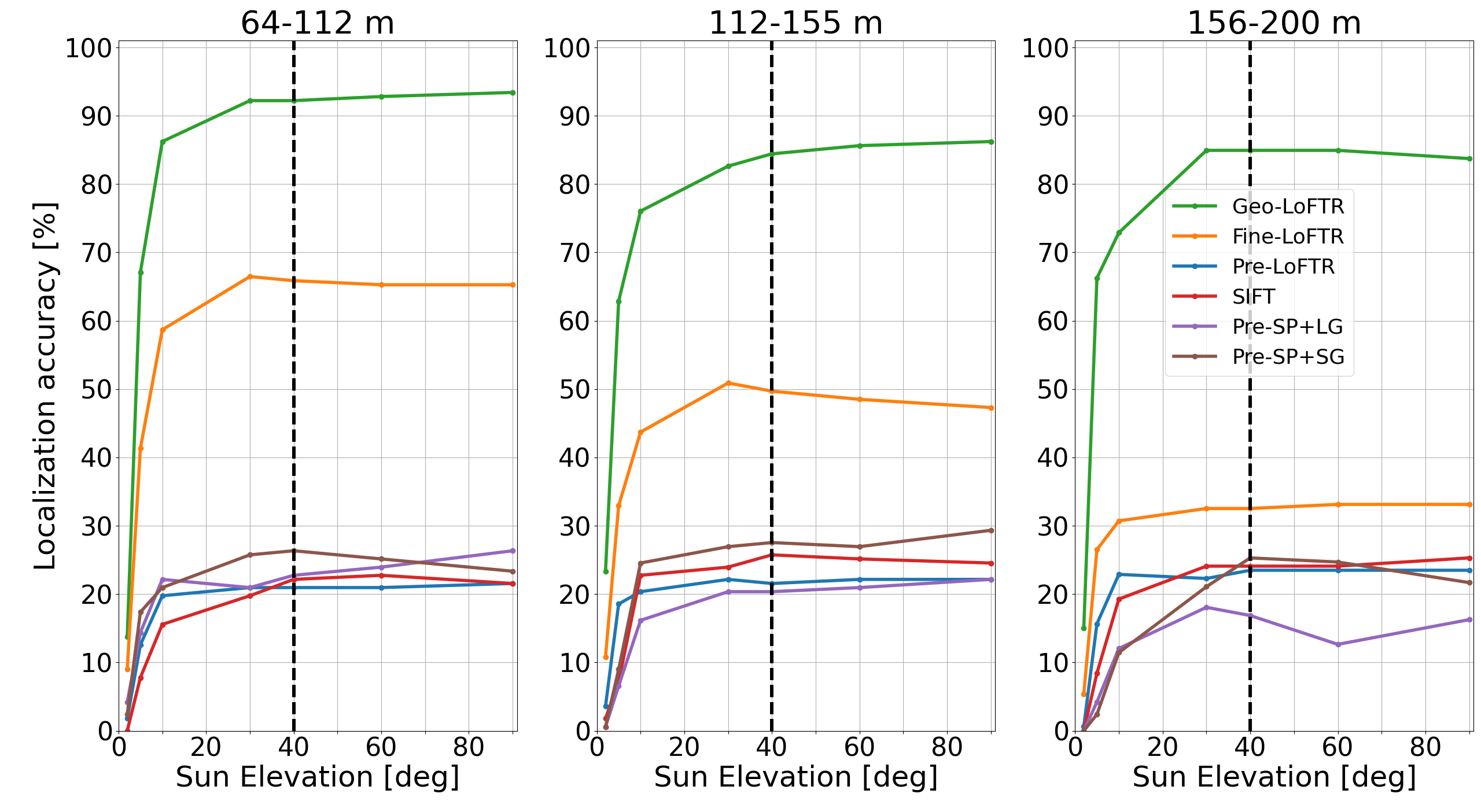}    
\end{minipage}
\\
\vspace{-2pt}
\begin{minipage}[b]{1.0\linewidth}
    \centering
    \includegraphics[width=1\linewidth]{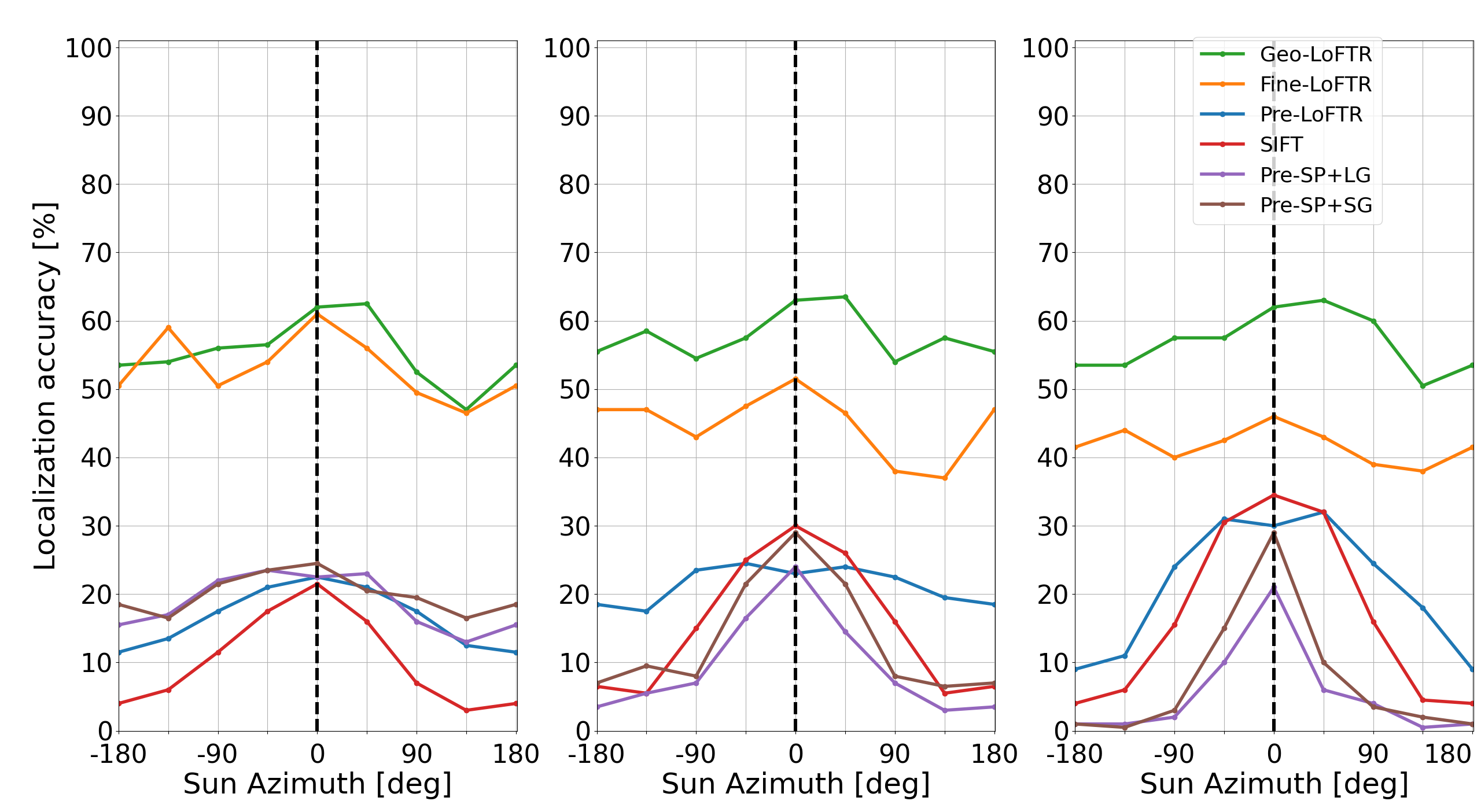}    
\end{minipage}
\vspace{-15pt}
\caption{Localization accuracy @1m as a function of map sun elevation (\textit{top}) and azimuth (\textit{bottom}) for test observations across three altitude ranges. Sun AZ angles are in the $[-180^{\circ}, 180^{\circ}]$ range. Map sun angles matching the observations are marked with a dashed line.}
\label{fig:mbl_loc_acc_vs_el_and_az_vs_alt}
\end{figure}

\noindent \textbf{Robustness to scale changes under varying lighting.}
We report the @1m accuracy as a function of map sun EL and AZ for observations taken within three different altitude ranges in Fig.~\ref{fig:mbl_loc_acc_vs_el_and_az_vs_alt}.
In contrast to Fine-LoFTR,  Geo-LoFTR maintains consistent localization performance over altitude variations. These results suggest that adding a geometric context might contribute to scale invariance, by leveraging the consistent pixel-to-pixel depth relationships across altitudes. 

\subsection{Localization Over a Simulated Martian Day}
\label{subsec:mbl_martian_day}
\begin{figure}
\setlength{\abovecaptionskip}{0pt}  
\setlength{\belowcaptionskip}{0pt}  
\centering
\vspace{-5pt}
\includegraphics[width=0.85\linewidth]{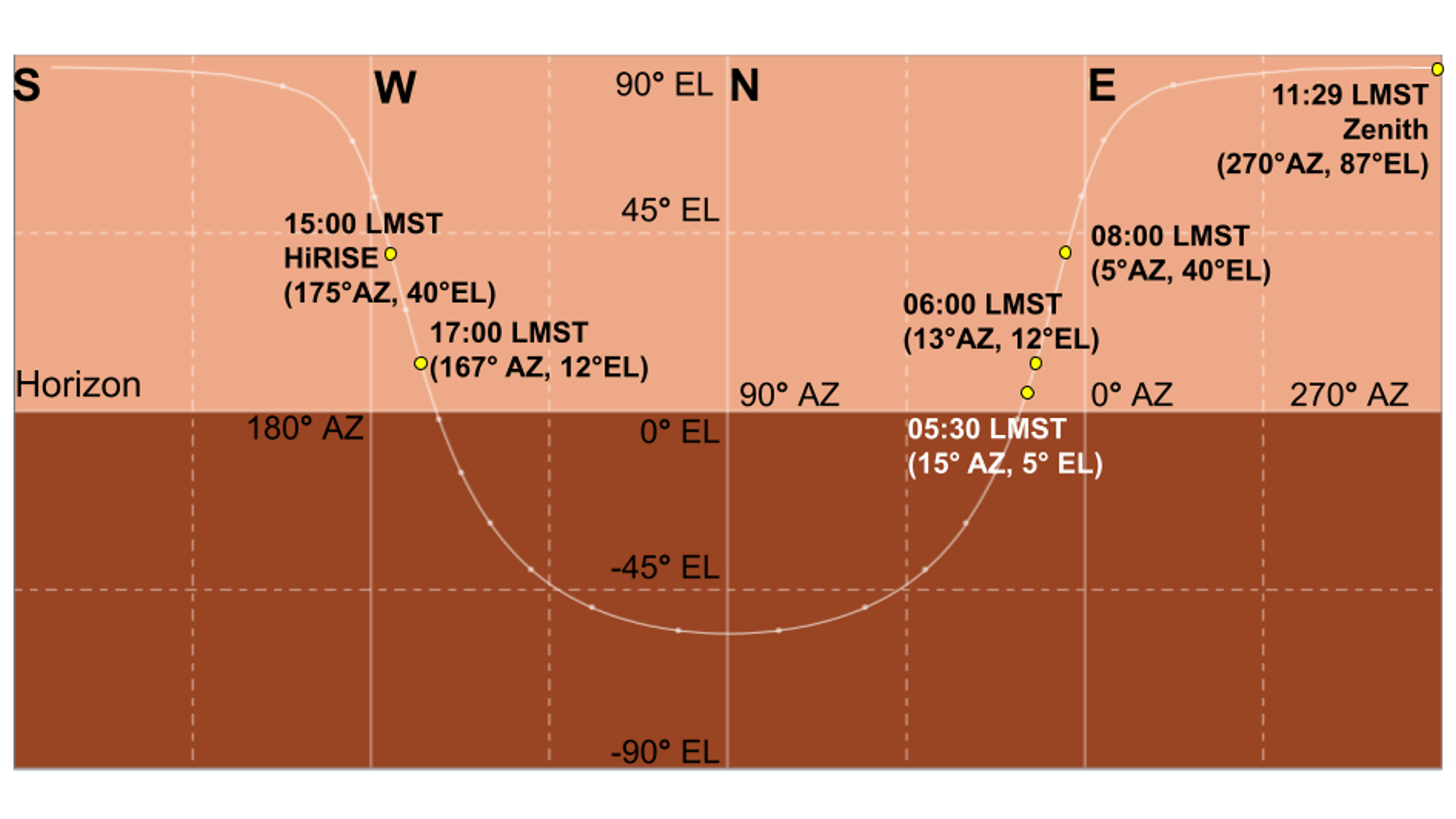}
\vspace{-5pt}
\caption{Sun trajectory on a local panorama from 77.44$^{\circ}$E longitude and 18.43$^{\circ}$N latitude on Mars, on 2031-05-10, with positions shown at four Local Mean Solar Times (LMSTs). Adapted from Mars24 \cite{mars24}.}
\label{fig:sun_profile}
\end{figure}
\begin{figure}
\centering
\makebox[0.01\linewidth]{}

\begin{minipage}[b]{0.01\linewidth}
    \raisebox{2em}[0pt][0pt]{\makebox[0pt][r]{\small \shortstack{LMST\\ 05:30}}} 
\end{minipage}
\begin{minipage}[b]{0.25\linewidth}
    \centering
    \includegraphics[width=\linewidth]{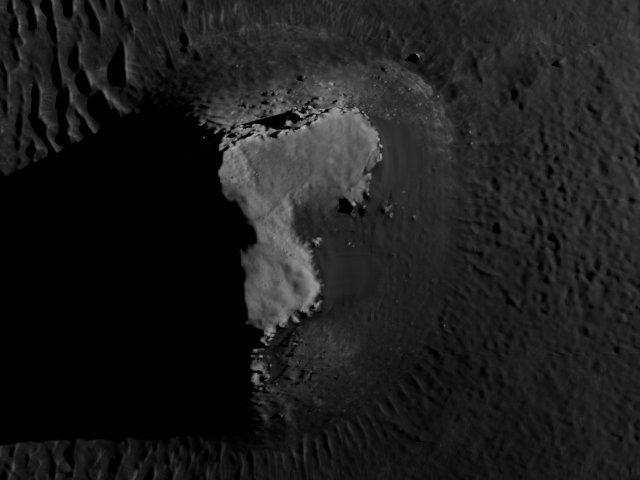}
\end{minipage}
\begin{minipage}[b]{0.25\linewidth}
    \centering
    \includegraphics[width=\linewidth]{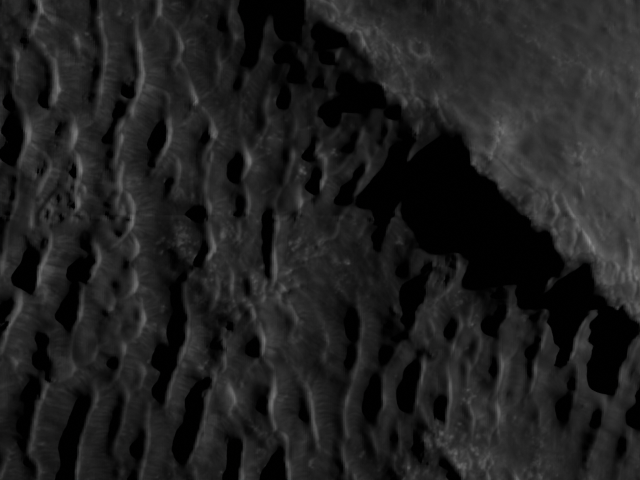}
\end{minipage}
\begin{minipage}[b]{0.25\linewidth}
    \centering
    \includegraphics[width=\linewidth]{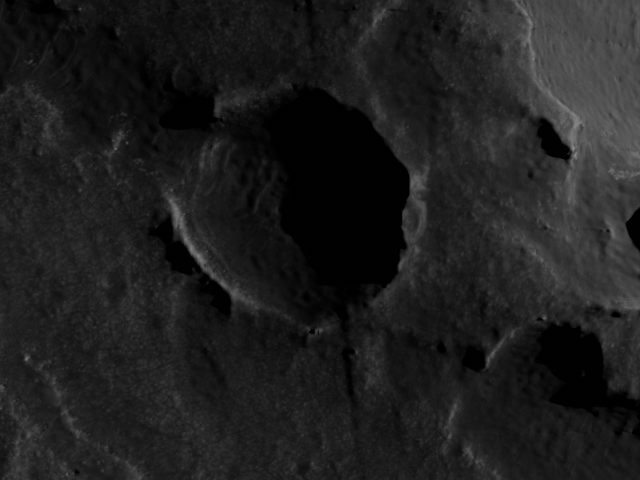}
\end{minipage}

\begin{minipage}[b]{0.01\linewidth}
    \raisebox{2em}[0pt][0pt]{\makebox[0pt][r]{\small 06:00
    }} 
\end{minipage}
\begin{minipage}[b]{0.25\linewidth}
    \centering
    \includegraphics[width=\linewidth]{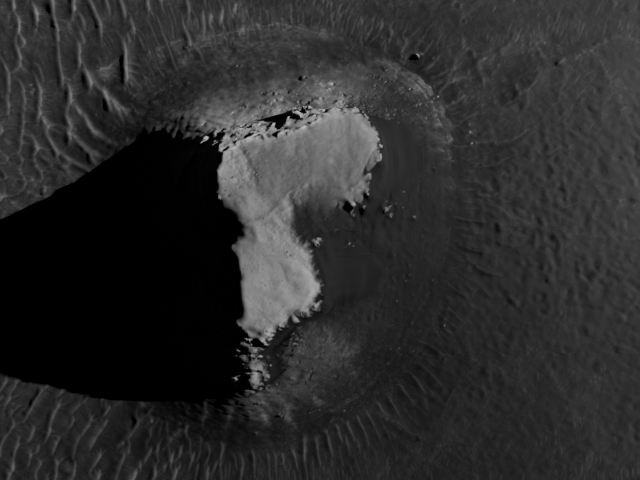}
\end{minipage}
\begin{minipage}[b]{0.25\linewidth}
    \centering
    \includegraphics[width=\linewidth]{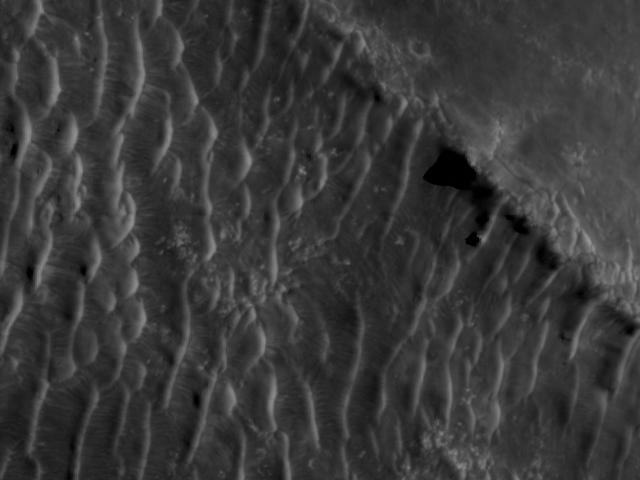}
\end{minipage}
\begin{minipage}[b]{0.25\linewidth}
    \centering
    \includegraphics[width=\linewidth]{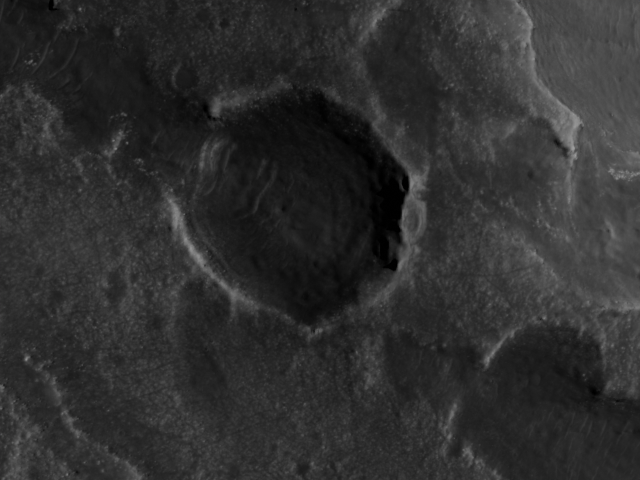}
\end{minipage}


\begin{minipage}[b]{0.01\linewidth}
    \raisebox{2em}[0pt][0pt]{\makebox[0pt][r]{\small11:29
    }} 
\end{minipage}
\begin{minipage}[b]{0.25\linewidth}
    \centering
    \includegraphics[width=\linewidth]{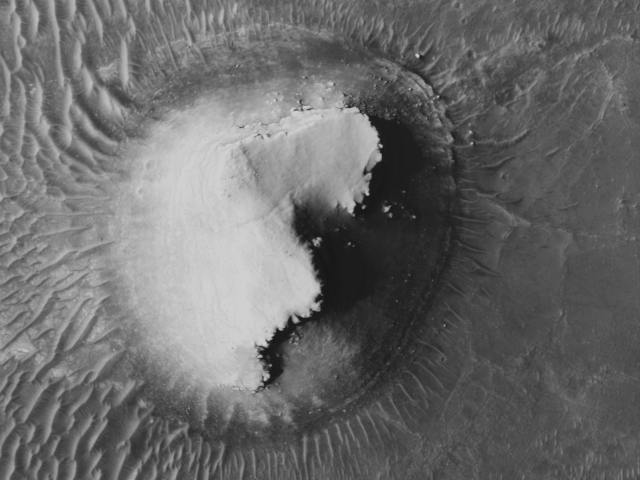}
\end{minipage}
\begin{minipage}[b]{0.25\linewidth}
    \centering
    \includegraphics[width=\linewidth]{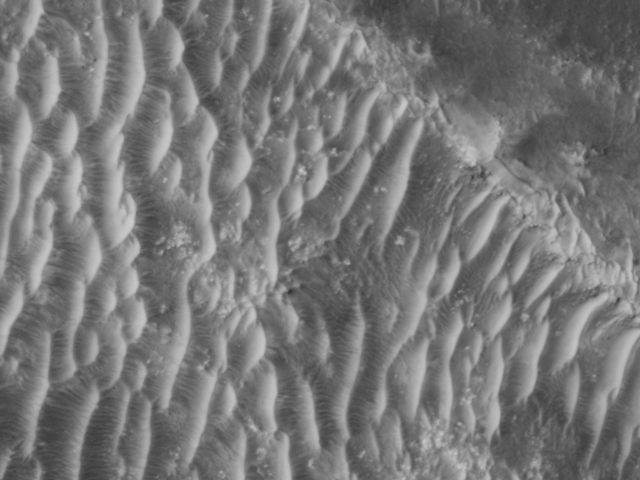}
\end{minipage}
\begin{minipage}[b]{0.25\linewidth}
    \centering
    \includegraphics[width=\linewidth]{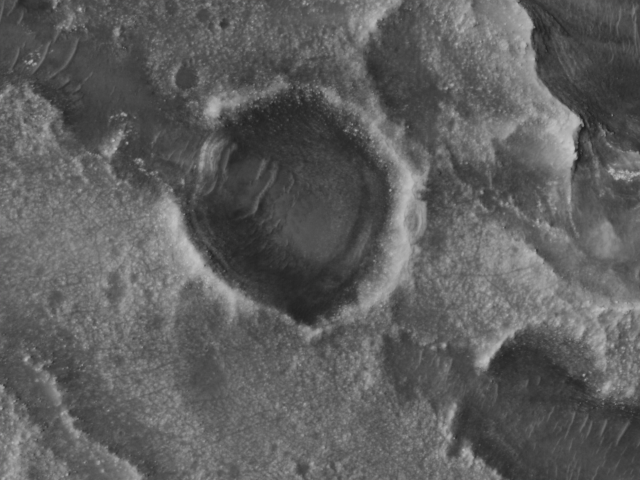}
\end{minipage}

\begin{minipage}[b]{0.01\linewidth}
    \raisebox{2em}[0pt][0pt]{\makebox[0pt][r]{\small15:00
    }} 
\end{minipage}
\begin{minipage}[b]{0.25\linewidth}
    \centering
    \includegraphics[width=\linewidth]{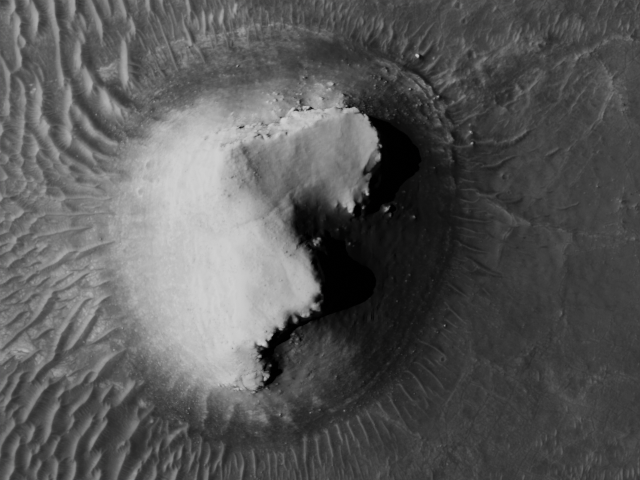}
\end{minipage}
\begin{minipage}[b]{0.25\linewidth}
    \centering
    \includegraphics[width=\linewidth]{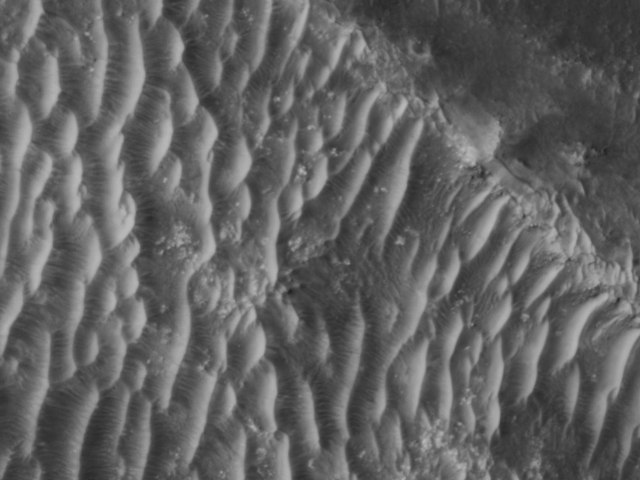}
\end{minipage}
\begin{minipage}[b]{0.25\linewidth}
    \centering
    \includegraphics[width=\linewidth]{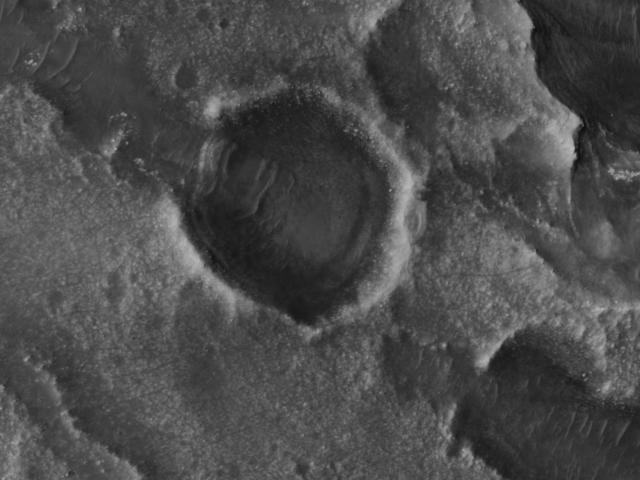}
\end{minipage}

\begin{minipage}[b]{0.01\linewidth}
    \raisebox{2em}[0pt][0pt]{\makebox[0pt][r]{\small17:00
    }} 
\end{minipage}
\begin{minipage}[b]{0.25\linewidth}
    \centering
    \includegraphics[width=\linewidth]{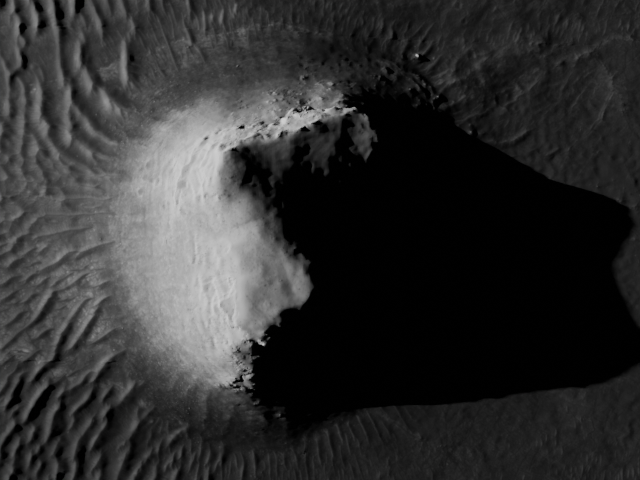}
\end{minipage}
\begin{minipage}[b]{0.25\linewidth}
    \centering
    \includegraphics[width=\linewidth]{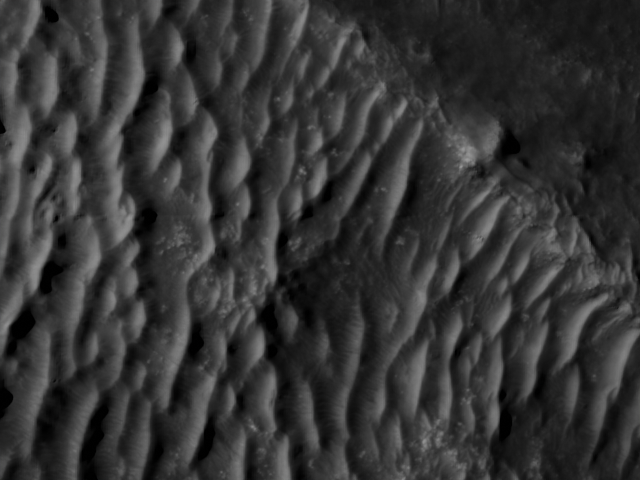}
\end{minipage}
\begin{minipage}[b]{0.25\linewidth}
    \centering
    \includegraphics[width=\linewidth]{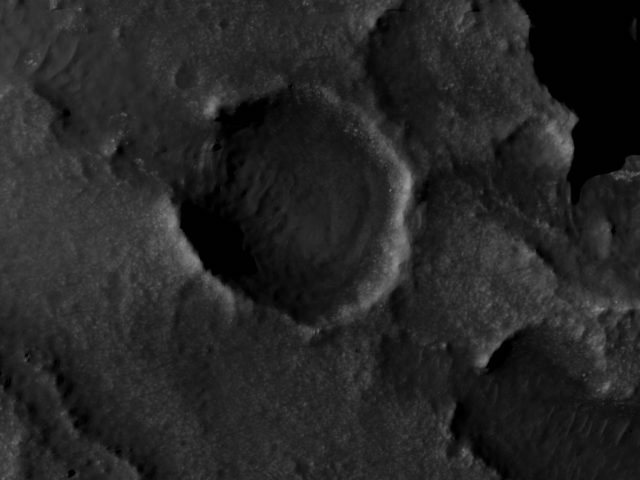}
\end{minipage}
\caption{Sample query observations rendered at different Local Mean Solar Times (LMSTs) taken on Mars on 2031-05-10. The reference HiRISE map is taken at 15:00 LMST. The sun Zenith is at 11:29 LMST.}
\label{fig:obsv_lmst}
\end{figure}
\begin{figure}
\centering
\vspace{3pt}
\includegraphics[width=1\linewidth]{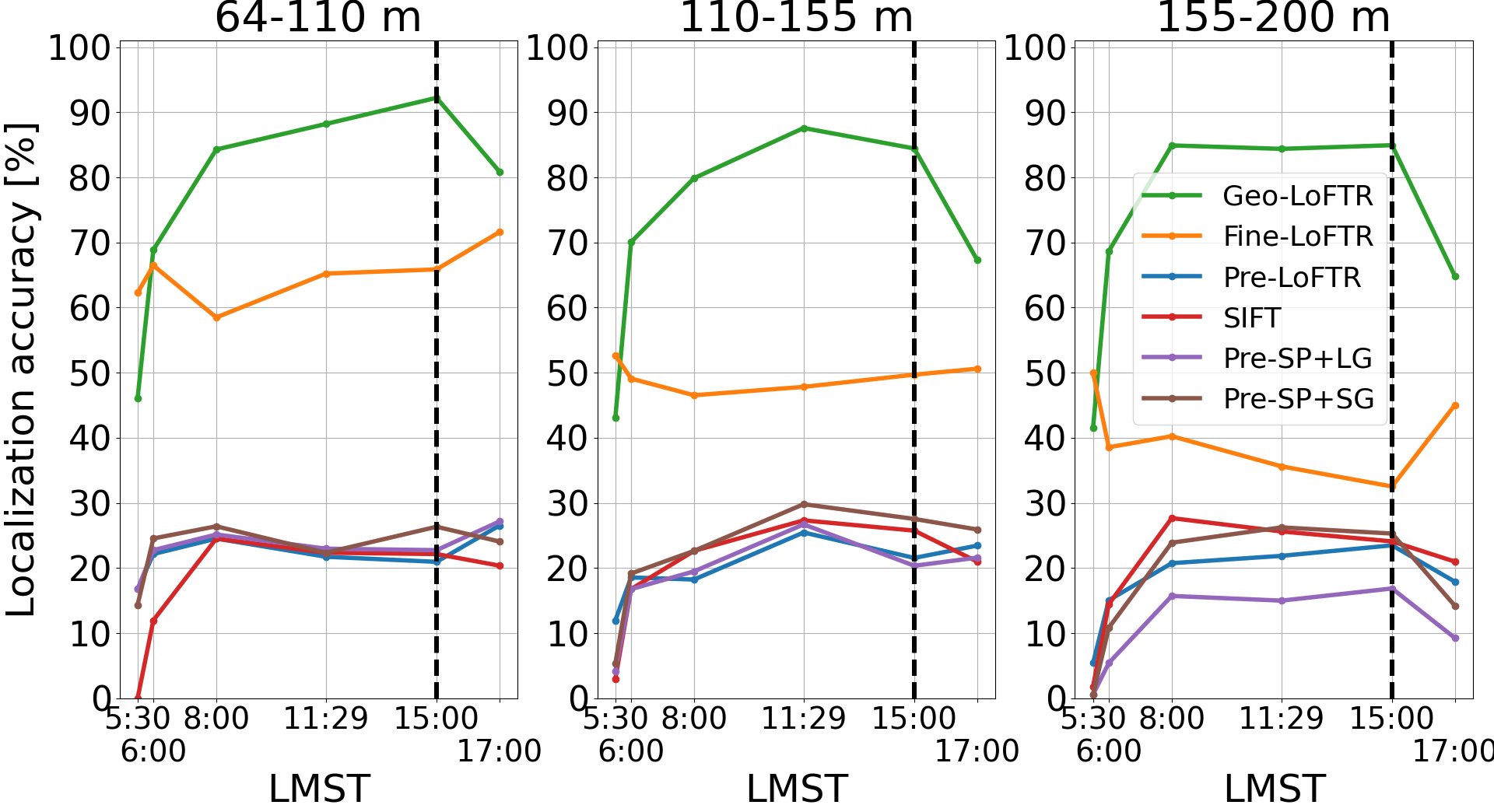}
\vspace{-18pt}
\caption{Localization accuracy (@1m) as a function of Local Mean Solar Time (LMST) of simulated test observations from the Jezero crater on 2031-05-10 across the 64-200 m altitude range. The reference HiRISE-like map is taken at 15:00 LMST (\textit{dashed line}). The sun Zenith is at 11:29 LMST.}
\label{fig:loc_acc_vs_lmst_1acc_vs_alt}
\end{figure}

The MbL performance is investigated for observations taken at different Local Mean Solar Times (LMSTs) during a simulated Martian day on the Jezero crater site.
We used the Mars24~\cite{mars24} software to compute the Sun's local trajectory for the selected date of 2031-05-10,
ensuring a broad range of sun elevations throughout the day (peak zenith 86.7$^{\circ}$ EL, Fig.~\ref{fig:sun_profile}). 
We generated nadir-pointing observations in MARTIAN at multiple times of the day from 5:30 to 17:00 LMST, totaling 3000 queries across the 64-200 m altitude range (Fig.~\ref{fig:obsv_lmst}). We also rendered an orthographic map at 15:00 LMST (175.1$^{\circ}$ AZ, 39.9$^{\circ}$ EL) as a HiRISE-like reference. 
Fig.~\ref{fig:loc_acc_vs_lmst_1acc_vs_alt} shows the @1m accuracy as function of LMSTs within three different altitude sub-ranges, with Geo-LoFTR being the most accurate for most of the Martian day.

\subsection{Registration of Mars2020 Descent Imagery}
\label{subsec:m2020_edl}
\begin{figure}
\centering
\includegraphics[width=0.7\linewidth]{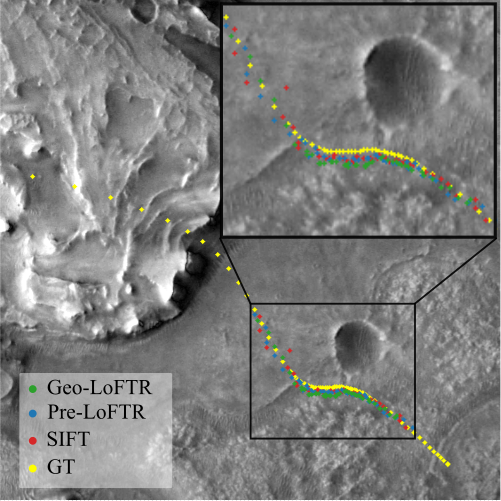}
\vspace{-3pt}
\caption{Reconstructed Mars2020 descent trajectory on a 9 km$^2$ crop of the CTX map of the Jezero crater, using localization estimates from Geo-LoFTR, Pre-LoFTR and SIFT in between 6 km and 960 m altitude. This range has been selected to match the observation-to-map scale ratio expected with HiRISE in the MSH flight envelop. GT: ground truth EDL trajectory.}
\label{fig:edl_loc_on_ctx}
\end{figure}

\begin{figure}
\vspace{3pt}
\centering
\includegraphics[width=0.9\linewidth]{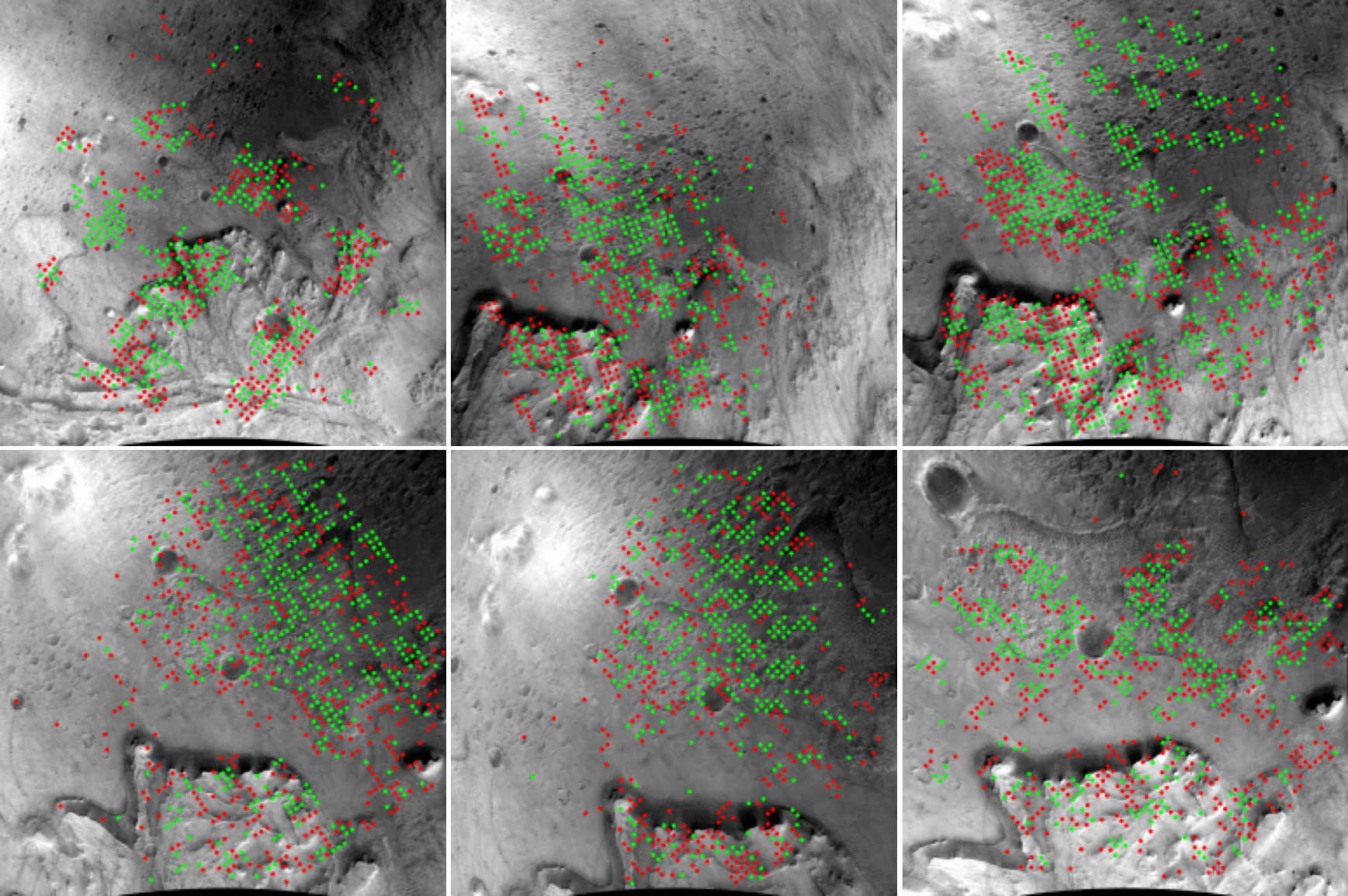} 
\vspace{-3pt}
\caption{Samples from the LCAM imaging sequence during the Mars2020 EDL overlaid with inlier (\textit{green}) and outlier (\textit{red}) correspondences produced by Geo-LoFTR with respect to the reference CTX map. Inliers retained after RANSAC-PnP (reprojection error $< 1$ px) indicate successful matching between real Mars descent imagery and the orbital map.}
\label{fig:edl_queries}
\end{figure}
In the absence of representative aerial imagery from Mars, we validate our method on actual Mars imagery by registering LVS camera (LCAM) images from the Mars 2020 EDL\footnote{https://mars.nasa.gov/mars2020/multimedia/raw-images/} to a CTX reference map with a resolution of 6 m/px. Although this does not capture the anticipated flight altitude of MSH, matching the descent frames between 6 km and 960 m to a CTX base map exhibits a similar scale change as for nominal MSH flights.
While illumination variations are absent
and the terrain geometry cues are largely diminished due to the high altitude, this experiment provides qualitative validation on actual Mars imagery.

The raw LCAM images (1024-px size) are processed in two steps. First, they are undistorted with a CAHVORE camera model~\cite{gennery2006generalized} and second, they are rotated in the image plane using the LCAM pose-prior estimates with respect to the reference map. The resulting images are finally registered onto a CTX map of 5000-px size. 
Fig~\ref{fig:edl_loc_on_ctx} shows the reconstructed descent trajectory on a map crop area of 9 $km^2$. 
Geo-LoFTR and Fine-LoFTR successfully provided global localization,
performing on par with the baselines as expected without relevant illumination changes. Examples of LCAM images are illustrated in Fig.~\ref{fig:edl_queries}, with Geo-LoFTR-matched keypoints. Rather than establishing accuracy bounds, this experiment assesses whether Geo-LoFTR, trained on synthetic data, can operate on real Mars imagery. The results indicate successful synthetic-to-real transfer under comparable scale changes. At the same time, the EDL sequence differs from the nominal orbital-to-UAV training regime, providing a limited indication of out-of-distribution generalization across flight altitudes. However, broader generalization across terrain types and map sources remains to be evaluated.

\subsection{Computational requirements}
\label{subsec:computational_req}
We report the LoFTR and Geo-LoFTR runtime as 543 ms $\pm$ 3 ms and 887 ms $\pm$ 27 ms respectively estimated over 100 forward passes on the NVIDIA embedded hardware Jetson AGX Xavier.
This benchmark was conducted using the plain Python implementations of the models, without acceleration frameworks, such as TensorRT, which are expected to significantly improve inference on edge devices. Computational efficiency optimization is left to future work.
We note that MbL does not need to run in real-time and its update frequency is dictated by mission requirements.
For example, assuming a maximum speed of 20 m/s, a maximum 5\% VIO position drift~\cite{Delaune_2021}, and an error tolerance of 5 m~\cite{anderson2024lessons}, then an MbL update would be needed once every 5 seconds. Furthermore, the 1km$^2$ search area adopted in our experiments simulates an extreme VIO failure scenario. In practice, it can be reduced given the reported VIO error~\cite{Delaune_2021}.

\section{CONCLUSION} 
\label{sec:conclusion}
In this paper, we presented a new map-based localization pipeline for future Mars helicopters that uses Geo-LoFTR, a geometry-aided feature matching model to register onboard images to reference maps. 
To generate a large-scale training dataset, we developed a custom simulation framework that uses real Digital Terrain Models from Mars.
Geo-LoFTR outperformed the baselines in terms of localization accuracy by a large margin, demonstrating that geometric context increases robustness to varying illumination and flight altitudes. Preliminary experiments on Mars2020 descent data demonstrate that Geo-LoFTR is applicable to real Martian imagery.
Future work will focus on optimizing Geo-LoFTR and the MbL pipeline for computational efficiency on high-performance edge devices (e.g. Jetson AGX Orin) for deployment on Mars surrogate UAVs during field tests on Earth.

\section*{ACKNOWLEDGMENTS}
The research described in this paper was carried out at the Jet Propulsion Laboratory, California Institute of Technology, under a contract with the National Aeronautics and Space Administration (80NM0018D0004). The authors  thank Andrew Johnson, Benjamin Morrell, and Shehryar Khattak for valuable discussions and feedback on this work.






\bibliographystyle{IEEEtran}
\bibliography{IEEEabrv, references}

\end{document}